\documentclass{article}

\usepackage{arxiv}

\usepackage[utf8]{inputenc} 
\usepackage[T1]{fontenc}    
\usepackage{hyperref}       
\usepackage{url}            
\usepackage{booktabs}       
\usepackage{amsfonts}       
\usepackage{nicefrac}       
\usepackage{microtype}      
\usepackage{lipsum}		
\usepackage{graphicx}
\usepackage{natbib}
\usepackage{doi}

\usepackage{wrapfig}
\usepackage{amsmath}
\usepackage{amssymb}
\usepackage{multirow}
\usepackage{float}
\usepackage{subfig}
\usepackage{amsthm}
\usepackage{algorithm}

\newtheorem{definition}{Definition}

\usepackage{amsmath,amsfonts,bm}

\def\eqref#1{equation~\ref{#1}}

\def\1{\bm{1}}

\DeclareMathAlphabet{\mathsfit}{\encodingdefault}{\sfdefault}{m}{sl}
\SetMathAlphabet{\mathsfit}{bold}{\encodingdefault}{\sfdefault}{bx}{n}

\usepackage{array}

\title{Enhanced Bayesian Optimization via Preferential Modeling of Abstract
	Properties}

\author{
	{Arun Kumar A V\thanks{Mail Correspondence: \texttt{a.anjanapuravenkatesh@deakin.edu.au}}$\:\:$,   Alistair Shilton, Sunil Gupta,  Santu Rana, Stewart Greenhill, Svetha Venkatesh}  \\
	Applied Artificial Intelligence Institute (A$^2$I$^2$)\\
	 Deakin University, Australia \\
}

\date{}

\renewcommand{\headeright}{}
\renewcommand{\undertitle}{}
\renewcommand{\shorttitle}{BOAP: Bayesian Optimization with Abstract Properties}

\hypersetup{
pdftitle={BOAP: Bayesian Optimization with Abstract Properties},
pdfsubject={BOAP},
pdfauthor={Arun Kumar A V},
pdfkeywords={Machine Learning, Bayesian Optimization, Gaussian Process, Human Expert, Preferential Modeling},
}

\begin{document}
\maketitle

\begin{abstract}
	Experimental (design) optimization is a key driver in
	designing and discovering new products and processes. Bayesian Optimization
	(BO) is an effective tool for optimizing expensive and black-box experimental
	design processes. While Bayesian optimization is a principled data-driven
	approach to experimental optimization, it learns everything from scratch
	and could greatly benefit from the expertise of its human (domain)
	experts who often reason about systems at different abstraction levels
	using physical properties that are not necessarily directly measured
	(or measurable). In this paper, we propose a human-AI collaborative
	Bayesian framework to incorporate expert preferences about unmeasured
	abstract properties into the surrogate modeling to further boost the
	performance of BO. We provide an efficient strategy that can also
	handle any incorrect/misleading expert bias in preferential judgments.
	We discuss the convergence behavior of our proposed framework. Our
	experimental results involving synthetic functions and real-world
	datasets show the superiority of our method against the baselines.
\end{abstract}


\section{Introduction\label{sec:introduction}}
Experimental design is the workhorse of scientific design and discovery.
Bayesian Optimization (BO) has emerged as a powerful methodology for
experimental design tasks \cite{martinez2014bayesopt,greenhill2020bayesian}
due to its sample-efficiency in optimizing expensive black-box functions.
In its basic form, BO starts with a set of randomly initialized designs
and then sequentially suggests the next design until the target objective
is reached or the optimization budget is depleted. Theoretical analyses
\cite{srinivas2012information,chowdhury2017kernelized} of BO methods
have provided mathematical guarantees of sample efficiency in the
form of sub-linear regret bounds. While BO is an efficient optimization
method, it only uses data gathered during the design optimization
process. However, in real world experimental design tasks, we also
have access to human experts \cite{swersky2017improving} who have
enormous knowledge about the underlying physical phenomena. Incorporating
such valuable knowledge can greatly accelerate the sample-efficiency
of BO.

Previous efforts in BO literature have incorporated expert knowledge
on the shape of functions \cite{kumar2019bayesian}, form of trends
\cite{licheng2018}, priors over optima \cite{hvarfner2022pi} and
model selection \cite{venkateshhuman}, which require experts to provide
very detailed knowledge about the black-box function. However, most
experts understand the process in an approximate or qualitative way,
and usually reason in terms of the intermediate abstract properties
- the expert will compare designs, and reason as to why one design
is better than another using high level abstractions. E.g, consider
the design of a spacecraft shield (Whipple shield) consisting of 2
plates separated by a gap to safeguard the spacecraft against micro-meteoroid
and orbital debris particle impacts. The design efficacy is measured
by observing the penetration caused by hyper-velocity debris. An expert
would reason why one design is better than another and accordingly
come up with a new design to try out. As part of their domain knowledge,
human experts often expect the first plate to shatter the space debris
while the second to absorb the fragments effect. Based on these abstract
intuitions, the expert will compare a pair of designs by examining
the shield penetration images and ask: Does the first plate shatter
better \emph{(Shattering)}? Does the second plate absorb the fragments
better \emph{(Absorption)}? The use of such abstractions allows experts
to predict the overall design objective thus resulting in an efficient
experimental design process. It is important to note that measuring
such abstractions is not usually feasible and only expert's qualitative
inputs are available. \emph{Incorporating such abstract properties
in BO for the acceleration of experimental design process is not well
explored}.

\begin{figure}
	\centering{}\includegraphics[width=0.55\columnwidth]{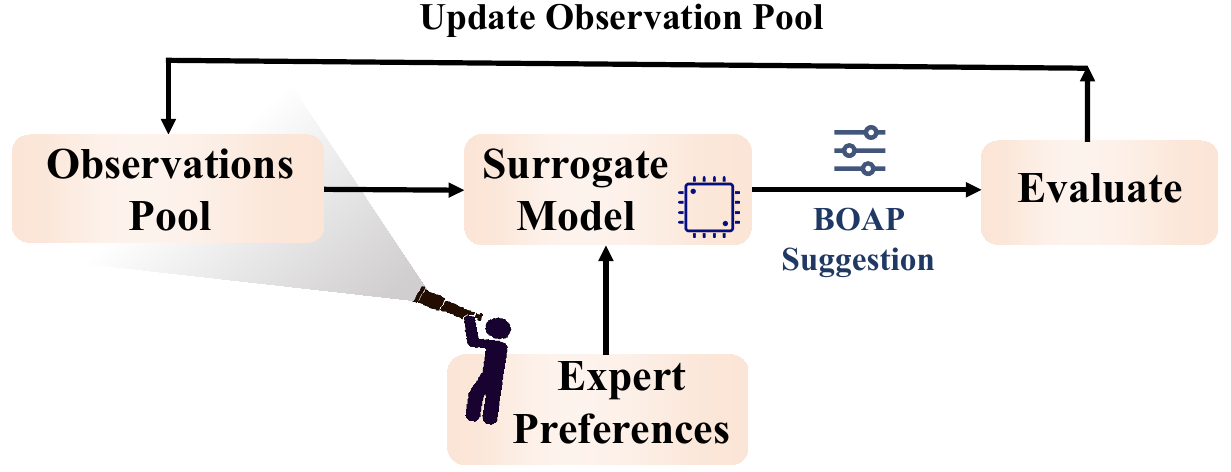}\caption{A schematic representation of Bayesian Optimization with Abstract
		Properties (BOAP)}
	\label{fig:schematic}\end{figure}

In this paper, we propose a novel human-AI collaborative approach
- \textbf{B}ayesian \textbf{O}ptimization with \textbf{A}bstract \textbf{P}roperties
(BOAP) - to accelerate BO by capturing expert inputs about the abstract,
unmeasurable properties of the designs. Since expert inputs are usually
qualitative \cite{nguyen2021top} and often available in the form
of design preferences based on abstract properties, we model each
abstract property via a latent function using the qualitative pairwise
rankings. We note that eliciting such pairwise preferences about designs
does not add significant cognitive overhead for the expert, in contrast
to asking for explicit knowledge about properties. We fit a separate
rank Gaussian process \cite{williams2006gaussian} to model each property.
Our framework allows enormous flexibility for expert collaborations
as it does not need the exact value of an abstract property, just
its ranking. A schematic of our proposed BOAP framework is shown in
Figure \ref{fig:schematic}.

Although we anticipate that experts will provide accurate preferences
on abstract properties, the expert preferential knowledge can sometimes
be misleading. Therefore to avoid such undesired bias, we use two
models for the black-box function. The first model uses a ``main''
Gaussian Process (GP) to model the black-box function in an augmented
input space where the design variables are augmented with the estimated
abstract properties modeled via their respective rank GPs. The second
model uses another ``main'' GP to model the black-box function using
the original design space \emph{without} any expert inputs. At each
iteration, we use predictive likelihood-based model selection to choose
the ``best'' model that has higher probability of finding the optima.

Our contributions are:\textbf{ (i)} we propose a novel human-AI collaborative
BO algorithm (BOAP) for incorporating the expert pairwise preferences
on abstract properties via rank GPs (Section \ref{sec:framework}),
\textbf{(ii)} we provide a theoretical discussion on the convergence
behavior of our proposed BOAP method (Section \ref{sec:convergence}),
\textbf{(iii)} we provide empirical results on both synthetic optimization
problems and real-world design optimization problems to prove the
usefulness of BOAP framework (Section \ref{sec:Experiments}).

\section{Background\label{sec:background}}
\textbf{}

\subsection{Bayesian Optimization}

Bayesian Optimization (BO) \cite{brochu2010tutorial,frazier2018tutorial}
provides an elegant framework for finding the global optima of an
expensive black-box function $f(\mathbf{x})$, given as{\small{} $\mathbf{x}^{\star}\in\text{argmax}_{\mathbf{x}\in\mathcal{X}}f(\mathbf{x})$,}
where $\mathcal{X}$ is a compact search space. BO is comprised of
two main components: (i) a surrogate model (usually a Gaussian Process
\cite{williams2006gaussian}) of the unknown objective function $f(\mathbf{x})$,
and (ii) an Acquisition Function $u(\mathbf{x})$ \cite{kushner1964new}
to guide the search for optima. 

\subsubsection{Gaussian Process}

A GP \cite{williams2006gaussian} is a flexible, non-parametric distribution
over functions. It is a preferred surrogate model because of its simplicity
and tractability, in contrast to other surrogate models such as Student-t
process \cite{shah2014student} and Wiener process \cite{weiner_process}.
A GP is defined by a prior mean function $\mu(\mathbf{x})$ and a
kernel $k:\mathcal{X}\times\mathcal{X}\text{\ensuremath{\rightarrow}}\mathbb{R}$.
The function $f(\mathbf{x}$) is modeled using a GP as {\small{}$f(\mathbf{x})\sim\mathcal{GP}(0,k(\mathbf{x},\mathbf{x}')).$
}If $\mathcal{D}_{1:t}=\{\mathbf{x}_{1:t},\mathbf{y}_{1:t}\}$ denotes
a set of observations, where $y=f(\mathbf{x})+\eta$ is the observation
corrupted with noise $\eta\in\mathcal{N}(0,\sigma_{\eta}^{2})$ then,
according to the properties of GP, the observed samples $\mathcal{D}_{1:t}$
and a new observation $(\mathbf{x}_{\star},f({\bf x}_{\star}))$ are
jointly Gaussian. Thus, the posterior distribution $\ensuremath{f({\bf x}_{\star})}$
is $\mathcal{N}(\mu(\mathbf{x}_{\star}),\sigma^{2}(\mathbf{x}_{\star}))$,
where $\mu(\mathbf{x}_{\star})=\mathbf{k}^{\intercal}[\mathbf{K}+\sigma_{\eta}^{2}\mathbf{I}]^{-1}\mathbf{y}_{1:t}$,
$\sigma^{2}(\mathbf{x}_{\star})=k(\mathbf{x}_{\star},\mathbf{x}_{\star})-\mathbf{k}^{\intercal}[\mathbf{K}+\sigma_{\eta}^{2}\mathbf{I}]^{-1}\mathbf{k}$,
$\mathbf{k=}[k(\mathbf{x}_{\star},\mathbf{x}_{1})\cdots k(\mathbf{x}_{\star},\mathbf{x}_{t})]^{\intercal}$,
and $\mathbf{K}=[k(\mathbf{x}_{i},\mathbf{x}_{j})]_{i,j\in\mathbb{N}_{t}}$.

\subsubsection{Acquisition Functions}

The acquisition function selects the next point for evaluation by
balancing the exploitation vs exploration (i.e searching in high value
regions vs highly uncertain regions). Some popular acquisition functions
include Expected Improvement \cite{Moc2}, GP-UCB \cite{srinivas2012information}
and Thompson sampling \cite{thompson1933likelihood}. A standard BO
algorithm is provided in Appendix $\S$ A.2.

\subsection{Rank GP Distributions}

\cite{kahneman2013prospect} demonstrated that humans are better at
providing qualitative comparisons than absolute magnitudes. Thus modeling
latent human preferences is crucial when optimization objectives in
domains such as A/B testing of web designing \cite{siroker2015b},
recommender systems \cite{brusilovski2007adaptive}, players skill
rating \cite{herbrich2006trueskill} and many more. \cite{chu2005preference}
proposed a non-parametric Bayesian algorithm for learning instance
or label preferences. We now discuss modeling pairwise preference
relations using rank GPs. 

Consider a set of $n$ distinct training instances denoted by $X=\{\mathbf{x}_{i}\:\forall i\in\mathbb{N}_{n}\:|\mathbb{N}_{n}=\{1,2,\cdots,n\}\}$
based on which pairwise preference relations are observed. Let $P=\{(\mathbf{x}\succ\mathbf{x}')\:|\;\mathbf{x},\mathbf{x}'\in X\}$
be a set of pairwise preference relations, where the notation $\mathbf{x}\succ\mathbf{x}'$
expresses the preference of instance $\mathbf{x}$ over $\mathbf{x}'$.
E.g., the pair $\{\mathbf{x},\mathbf{x}'\}$ can be two different
spacecraft shield designs and $\mathbf{x}\succ\mathbf{x}'$ implies
that spacecraft design $\mathbf{x}$ is preferred over $\mathbf{x}'$.
\cite{chu2005preference} assume that each training instance is associated
with an unobservable latent function value $\{\bar{f}(\mathbf{x})\}$
measured from an underlying hidden preference function $\bar{f}:\mathbb{R}^{d}\rightarrow\mathbb{R}$,
where $\mathbf{x}\succ\mathbf{x}'$, implies $\bar{f}(\mathbf{x})>\bar{f}(\mathbf{x}')$.
Employing an appropriate GP prior and likelihood, user preference
can be modeled via rank GPs.

Preference learning has been used in BO literature \cite{gonzalez2017preferential,mikkola2020projective}.
\cite{gonzalez2017preferential} proposed Preferential BO (PBO) to
model the unobserved objective function using a binary design preferential
feedback. \cite{benavoli2021preferential} modified PBO to compute
posteriors via skew GPs. \cite{astudillo2020multi} proposed a preference
learning based BO to model preferences in a multi-objective setup
using multi-output GPs. All these works incorporate preferences about
an unobserved objective function. However, in this paper, we use preference
learning to model expert preferences about the intermediate abstract
(auxiliary) properties\textit{. }\textit{\emph{Our latent model learnt
using such preferential data is then used as an input to model the
main objective function.}}

\section{Framework\label{sec:framework}}
This paper addresses the global optimization of an expensive, black-box
function,\textit{ i.e} we aim to find:\textbf{
\begin{equation}
\mathbf{x}^{\star}\in\underset{\textbf{\ensuremath{\mathbf{x}}\ensuremath{\in}}\mathcal{X}}{\operatorname{argmax}f(\mathbf{x})}\label{eq:bo_opt_prob}
\end{equation}
}where $f:\mathcal{X}\rightarrow\mathbb{R}$ is a noisy and expensive
objective function. For example, $f$ could be a metric signifying
the strength of the spacecraft design. The motivation of this research
work is to model $f$ by capturing the cognitive knowledge of experts
in making preferential decisions based on the inherent non-measurable
abstract properties of the possible designs. The objective here is
same as that of standard BO \emph{i.e., }to find the optimal design
($\mathbf{x}^{\star}$) that maximizes the unknown function $f$,
but in the light of expert preferential knowledge on abstract properties.
The central idea is to use preferential feedback to model and utilize
the underlying higher-order properties that underpin preferential
decisions about designs. We propose \textbf{B}ayesian \textbf{O}ptimization
with \textbf{A}bstract \textbf{P}roperties (\textbf{BOAP}) for the
optimization of $f$ in the light of expert preferential inputs. First,
we discuss expert knowledge about abstract properties. Next, we discuss
GP modeling of $f$ with preferential inputs, followed by a model-selection
step that is capable of overcoming a futile expert bias in preferential
knowledge. A complete algorithm for BOAP is presented in Algorithm
\ref{alg:boap} at the end of this section. 

\subsection{Expert Preferential Inputs on Abstract Properties \label{subsec:Expert-Preferential-Inputs}}

In numerous scenarios, domain experts reason the output of a system
in terms of higher order properties $\omega_{1}(\mathbf{x}),\omega_{2}(\mathbf{x}),\dots$
of a design $\mathbf{x}\in\mathcal{X}$. However, these abstract properties
are rarely measured, only being accessible via expert preferential
inputs. E.g., a material scientist designing spacecraft shield can
easily provide her pairwise preferences on the properties such as
shattering, shock absorption, \emph{i.e.,} \textit{``this design
absorbs shock better than that design''}, in contrast to specifying
the exact measurements of shock absorption. These properties can be
simple physical properties or abstract combinations of multiple physical
properties which an expert uses to reason about the output of a system.
We propose to incorporate such qualitative properties accessible to
the expert in the surrogate modeling of the given objective function
to further accelerate the sample-efficiency of BO. 

Let $\omega_{1:m}(\mathbf{x})$ be a set of $m$ abstract properties
derived from the design $\mathbf{x}\in\mathcal{X}$. For property
$\omega_{i}$, design $\mathbf{x}$ is preferred over design $\mathbf{\ensuremath{x}}'$
if $\omega_{i}(\mathbf{x})>\omega_{i}(\mathbf{x}')$. \textit{\emph{We
denote the set of preferences provided on }}$\omega_{i}$\textit{\emph{
as }}$P^{\omega_{i}}=\{(\mathbf{x}\succ\mathbf{x}')\:\text{if}\:\omega_{i}(\mathbf{x})>\omega_{i}(\mathbf{x}')\:|\;\mathbf{x}\in\mathcal{X}\}$.

\subsubsection{Rank GPs for Abstract Properties}

We capture the aforementioned expert preferential data for each of
the abstract properties $\omega_{1:m}$ individually using $m$ separate
rank Gaussian process distributions \cite{chu2005preference}. In
conventional GPs the observation model consists of a map of input-output
pairs. In contrast, the observation model of rank (preferential) Gaussian
Process ($\mathcal{GP}$) consists of a set of instances and a set
of pairwise preferences between those instances. The central idea
here is to capture the ordering over instances $X=\{\mathbf{x}_{i}\:|\:\forall i\in\mathbb{N}_{n}\}$
by learning latent preference functions $\{\omega_{i}\:|\:\forall i\in\mathbb{N}_{m}\}$.
We denote such a rank GP modeling abstract property $\omega_{i}$
by the notation $\mathcal{GP}_{\omega_{i}}$.

Let $\mathcal{X}\in\mathbb{R}^{d}$ be a $d-$dimensional compact
search space and $X=\{\mathbf{x}_{i}\:|\:\forall i\in\mathbb{N}_{n}\}$
be a set of $n$ training instances. Let $\boldsymbol{\omega}=\{\omega(\mathbf{x})\}$
be the unobservable latent preference function values associated with
each of the instances $\mathbf{x}\in X$. Let $P$ be the set of $p$
pairwise preferences between instances in $X$, defined as $P=\{(\mathbf{x}\succ\mathbf{x}')_{j}\:\text{if}\:\omega(\mathbf{x})>\omega(\mathbf{x}')\:|\;\mathbf{x}\in X,\forall j\in\mathbb{N}_{p}\}$,
where $\omega$ is the latent preference function. The observation
model for the rank GP distribution $\mathcal{GP}_{\omega}$ modeling
the latent preference function $\omega$ is given as $\mathcal{\bar{D}}=\{\mathbf{x}_{1:n},P=\{(\mathbf{x}\succ\mathbf{x}')_{j}\:\forall\mathbf{x},\mathbf{x}'\in X,j\in\mathbb{N}_{p}\}\}$.

We follow the probabilistic kernel approach for preference learning
\cite{chu2005preference} to formulate the likelihood function and
Bayesian probabilities. Imposing non-parametric GP priors on the latent
function values $\mathbf{\boldsymbol{\omega}}$, we arrive at the
prior probability of $\boldsymbol{\omega}$ given by:

\begin{equation}
\mathcal{P}(\mathbf{\boldsymbol{\omega}})=(2\pi)^{-\frac{n}{2}}|\mathbf{K}|^{-\frac{1}{2}}\exp\big({\scriptstyle -\frac{1}{2}}\mathbf{\boldsymbol{\omega}}^{\intercal}\mathbf{K}^{-1}\mathbf{\boldsymbol{\omega}}\big)\label{eq:prior_prob_fbar}
\end{equation}

With suitable noise assumptions $\mathcal{N}(0,\tilde{\sigma}_{\eta}^{2})$
on inputs and the preference relations $(\mathbf{x},\mathbf{x}')_{1:m}$
in $P$, the Gaussian likelihood function based on \cite{thurstone2017law}
is:

\begin{equation}
\mathcal{P}(\mathbf{(\mathbf{x}\succ\mathbf{x}')}_{i}|\omega(\mathbf{x}),\omega(\mathbf{x}'))=\Phi\big(z_{i}(\mathbf{x},\mathbf{x}')\big)\label{eq:likelihood_function}
\end{equation}
where $\Phi$ is the c.d.f of standard normal distribution $\mathcal{N}(0,1)$
and $z(\mathbf{x},\mathbf{x}')=\frac{\omega(\mathbf{x})-\omega(\mathbf{x}')}{\sqrt{2\tilde{\sigma}_{\eta}^{2}}}$.
Based on Bayes theorem, the posterior distribution of the latent function
given the data is given by: 

\[
\mathcal{P}(\boldsymbol{\omega}|\mathcal{\bar{D}})=\frac{\mathcal{P}(\boldsymbol{\omega})}{\mathcal{P}(\mathcal{\bar{D}})}\mathcal{P}(\mathcal{\bar{D}}|\boldsymbol{\omega})
\]
where $\mathcal{P}(\mathbf{\boldsymbol{\omega}})$ is the prior distribution
(Eq. (\ref{eq:prior_prob_fbar})), $\mathcal{P}(\mathcal{\bar{D}}|\boldsymbol{\omega})$
is the probability of observing the pairwise preferences given the
latent function values $\boldsymbol{\omega}$, which can be computed
as a product of the likelihood (Eq. (\ref{eq:likelihood_function}))
\emph{i.e.,} $\mathcal{P}(\mathcal{\bar{D}}|\mathbf{\boldsymbol{\omega}})=\prod_{p}\mathcal{P}(\mathbf{(\mathbf{x}\succ\mathbf{x}')}_{p}|\omega(\mathbf{x}),\omega(\mathbf{x}'))$
and $\mathcal{P}(\mathcal{\bar{D}})=\int\mathcal{P}(\mathcal{\bar{D}}|\boldsymbol{\omega})\mathcal{P}(\boldsymbol{\omega})\:d\boldsymbol{\omega}$
is the evidence of model parameters including kernel hyperparameters.
We find the posterior distribution using Laplace approximation and
the Maximum A Posteriori estimate (MAP) $\boldsymbol{\omega}_{\text{MAP}}$
as the mode of posterior distribution. We can find the MAP using Newton-Raphson
descent given by:
\begin{equation}
\boldsymbol{\omega}^{\text{new}}=\boldsymbol{\omega}^{\text{old}}-\mathbf{H}^{-1}\mathbf{g}|_{\boldsymbol{\omega}=\boldsymbol{\omega}^{\text{old}}}\label{eq:newton_raphson_map}
\end{equation}
where the Hessian $\mathbf{H}=[\mathbf{K}+\tilde{\sigma}_{\eta}^{2}\mathbf{I}]^{-1}+\mathbf{C}$,
and the gradient $\mathbf{g}=\nabla_{\boldsymbol{\omega}}\log\:\mathcal{P}(\boldsymbol{\omega}|\mathcal{\bar{D}})=-[\mathbf{K}+\tilde{\sigma}_{\eta}^{2}\mathbf{I}]^{-1}\boldsymbol{\omega}+\mathbf{b}$,
given {\footnotesize{}$b_{j}=\frac{\partial}{\partial\omega(\mathbf{x}_{j})}\sum\limits _{p}\ln\Phi(z_{p})$
and $C_{ij}=\frac{\partial}{\partial\omega(\mathbf{x}_{i})\partial\omega(\mathbf{x}_{j})}\sum\limits _{p}\ln\Phi(z_{p})$.}{\footnotesize\par}

\subsubsection{Hyperparameter Optimization }

Kernel hyperparameter ($\theta^{\star}$) is crucial to optimize the
generalization performance of the GP. We perform the model selection
for our rank-GPs by maximizing the corresponding log-likelihood in
the light of latent values $\boldsymbol{\omega}_{\text{MAP}}$. In
contrast to the evidence maximization mentioned in \cite{chu2005preference}
\emph{i.e.,} $\theta^{\star}=\text{argmax}_{\theta}\:\mathcal{P}(\mathcal{\bar{D}}|\theta)$,
we find the optimal kernel hyperparameters by maximizing the log-likelihood
($\mathcal{\bar{L}}$) of rank GPs \emph{i.e.,} $\theta^{\star}=\text{argmax}_{\theta}\,\mathcal{\bar{L}}$.
The closed-form of log-likelihood of the rank GP is given as:

\vspace{-0.4cm}
{\footnotesize{}
\begin{equation}
\!\!\!\!\mathcal{\bar{L}}\!=\!-\frac{1}{2}\textbf{\ensuremath{\boldsymbol{\omega}_{\text{MAP}}^{\intercal}}}[\mathbf{K}\!+\tilde{\sigma}_{\eta}^{2}\mathbf{I}]^{{\scriptscriptstyle {\scriptscriptstyle -1}}}\boldsymbol{\omega}_{\text{MAP}}-\!\frac{1}{2}\!\log\!\!\mid\!\!\mathbf{K}\!+\tilde{\sigma}_{\eta}^{2}\mathbf{I}\!\mid\!\!-\frac{n}{2}\!\log(2\pi)\!\label{eq:closed_gp_log_likelihood}
\end{equation}
}{\footnotesize\par}

\subsection{Augmented GP with Abstract Property Preferences \label{subsec:Augmented-GP}}

To account for property preferences in modeling $f$, we augment the
input $\mathbf{x}$ of a conventional GP modeling $f$ with the mean
predictions obtained from $m$ rank GPs ($\mathcal{GP}_{\omega_{1:m}}$)
as auxiliary inputs capturing the property preferences $\omega_{1:m}$,
\textit{i.e.,} instead of modeling GP directly on $\mathbf{x}$ we
model on $\tilde{\mathbf{x}}=[\mathbf{x},\mu_{\omega_{1}}(\mathbf{x}),\cdots,\mu_{\omega_{m}}(\mathbf{x})]$,
where $\mu_{\omega_{i}}$ is the predictive mean computed using:
\begin{center}
$\mu_{\omega_{i}}(\mathbf{x})=\mathbf{k}^{\intercal}[\mathbf{K+}\sigma_{\eta}^{2}\mathbf{I}]^{-1}\boldsymbol{\omega}_{\text{MAP}}$
\par\end{center}

where $\mathbf{k}=[k(\mathbf{x},\mathbf{x}_{1}),\cdots,k(\mathbf{x},\mathbf{x}_{n})]^{\intercal},$
$\mathbf{K}=[k(\mathbf{x}_{i},\mathbf{x}_{j})]_{i,j\in\mathbb{N}_{n}}$
and $\mathbf{\mathbf{x}}_{i}\in X$. To handle different scaling levels
in rank GPs, we normalize its output in the interval $[0,1]$, such
that $\mu_{\omega_{i}}(\mathbf{x})\in[0,1]$.

Although we model $\tilde{\mathbf{x}}$ using mean predictions $\mu_{\omega_{i}}(\mathbf{x})$,
the uncertainty estimates were not (directly) considered in the modeling.
The GP predictive variance tends to be high outside of the neighborhood
of observations, indicating the uncertainty in our beliefs on the
model. Therefore, a data point with high predictive variance $(\sigma_{\omega_{1}}(\mathbf{x}))^{2}$
in rank GP indicate the model uncertainty. We incorporate this uncertainty
in our main GP modeling $\tilde{\mathbf{x}}$ such that the effects
of predicted abstract properties $\mu_{\omega_{i}}(\mathbf{x})$ are
appropriately reduced when the model is uncertain i.e. when $(\sigma_{\omega_{i}}(\mathbf{x}))^{2}$
is high. 

To achieve this, we formulate the feature-wise lengthscales as a function
of predictive uncertainty of the augmented dimensions to control their
\textit{importance} in the overall GP. Note that augmented features
can be detrimental when the model is uncertain. To address this potential
problem, we use a spatially varying kernel \cite{kumar2019bayesian}
that treats the lengthscale as a function of the input, rather than
a constant. A positive definite kernel with spatially varying lengthscale
is given as: 

\begin{equation}
k(\mathbf{x},\mathbf{x}')=\prod_{d=1}^{D}\sqrt{\frac{2l(x_{d})l(x'_{d})}{l^{2}(x_{d})+l^{2}(x'_{d})}}\:\exp\bigg(-\sum_{d=1}^{D}\frac{(x_{d}-x'_{d})^{2}}{l^{2}(x_{d})+l^{2}(x'_{d})}\bigg)\label{eq:spatial_kernel}
\end{equation}
where $l(\cdot)$ is the lengthscale function. In our proposed framework,
we use lengthscale as a function i.e., $l(\cdot)$ only for the newly
augmented dimensions and set the lengthscale of the original dimensions
to a constant value i.e. $\theta=[l_{1},\cdots,l_{D},l_{\omega_{1}}(\mathbf{x}),\cdots,l_{\omega_{m}}(\mathbf{x})]$.
As we need lengthscale function to reflect the model uncertainty,
we set $l_{\omega_{i}}(\mathbf{x})=\alpha\tilde{\sigma}_{\omega_{i}}(\mathbf{x})$,
where $\tilde{\sigma}_{\omega_{i}}(\mathbf{x})$ is the normalized
standard deviation of the rank GP predicted for the abstract property
$\omega_{i}$ and $\alpha$ is a scale parameter that is tuned using
the standard GP log-marginal likelihood in conjunction with other
kernel parameters. The aforementioned lengthscales ensure that the
data points $\tilde{\mathbf{x}}$ with high model uncertainty have
higher lengthscale on the augmented dimensions and thus are treated
as less important.

The objective function is modeled on the concatenated inputs $\tilde{\mathbf{x}}\in\mathbb{R}^{d+m}$
and we denote this function with augmented inputs $\tilde{\mathbf{x}}$
as human-inspired objective function $h(\mathbf{\tilde{x}})$. The
GP ($\mathcal{GP}_{h}$) constructed in the light of expert preferential
data is then used in BO to find the global optima of $h(\mathbf{\tilde{x}})$,
given as:
\begin{equation}
\mathbf{x}^{\star}\in\underset{\mathbf{\mathbf{x}\in\mathcal{X}}}{\operatorname{argmax}}\,h(\mathbf{\tilde{x}})\label{eq:human_opt_prob}
\end{equation}
The observation model is $\mathcal{D}=\{(\mathbf{x},y=h(\tilde{\mathbf{x}})\approx f(\mathbf{x}))\}$\emph{
i.e., }the human-inspired objective function $h(\tilde{\mathbf{x}})$
is a simplified $f(\mathbf{x})$ with auxiliary features in the input,
thus we observe the $h(\tilde{\mathbf{x}})$ via $f(\mathbf{x})$.

\subsection{Overcoming Inaccurate Expert Inputs}

Up to this point we have assumed that expert input is accurate and
thus likely to accelerate BO. However, in some cases this feedback
may be inaccurate, and potentially slowing optimization. To overcome
such bias and encourage exploration we maintain $2$ models, one of
which is augmented by expert abstract properties (we refer to this
as Human Arm-$\mathfrak{h}$) and an un-augmented model (we refer
to this as Control Arm-$\mathfrak{f}$), and use predictive likelihood
to select the arm at each iteration. 

The experimental arm models $f$ directly by observing the function
values at suggested candidate points. Here, we fit a standard GP ($\mathcal{GP}_{f}$)
based on the data collected \emph{i.e.,} $\mathcal{D}=\{(\mathbf{x},y=f(\mathbf{x})+\eta)\}$
where $\eta\sim\mathcal{N}(0,\sigma_{\eta}^{2})$ is the Gaussian
noise. The GP distribution ($\mathcal{GP}_{f})$ may be used to optimize
$f$ using a BO algorithm with Thompson Sampling (TS) \cite{thompson1933likelihood}
strategy. 

At each iteration $t$, we compare the predictive likelihoods ($\mathcal{L}_{t}$)
of both the human augmented arm (Arm-$\mathfrak{h}$) and an experimental
arm (Arm-$\mathfrak{f}$) to select the arm to pull for suggesting
the next promising candidate for the function evaluation. Then, we
use Thompson sampling to draw a sample $S_{t}$ from the GP posterior
distribution of the arm pulled and find its corresponding maxima given
as:
\begin{center}
{\small{}$\mathbf{x}_{t}^{\mathfrak{h}}=\text{\ensuremath{\underset{\mathbf{x}\in\mathcal{X}}{\operatorname{argmax}}\,}}(S^{\text{\ensuremath{\mathfrak{h}}}}(\tilde{\mathbf{x}}));\quad\mathbf{x}_{t}^{\mathfrak{f}}=\text{\ensuremath{\underset{\mathbf{x}\in\mathcal{X}}{\operatorname{argmax}}\,}}(S^{\text{\ensuremath{\mathfrak{f}}}}(\mathbf{x}))$ }{\small\par}
\par\end{center}

The arm with maximum predictive likelihood is chosen at each iteration
and we observe $f$ at the suggested location \textit{i.e.,} $(\mathbf{x}_{t}^{\mathfrak{h}},f(\mathbf{x}_{t}^{\mathfrak{h}}))$
or $(\mathbf{x}_{t}^{\mathfrak{f}},f(\mathbf{x}_{t}^{\mathfrak{f}}))$.
Then rank GPs are updated to capture the preferences with respect
to the new suggestion $\mathbf{x}_{t}^{\mathfrak{h}}$ or $\mathbf{x}_{t}^{\mathfrak{f}}$.
This process continues until the evaluation budget $T$ is exhausted.
Additional details of BOAP framework is provided in the Appendix ($\S$
A.3). 

\begin{algorithm}
\caption{BO with Preferences on Abstract Properties (BOAP)}
\label{alg:boap}

\textbf{Input}: Initial Observations: $\mathcal{D}_{1:t'}=\{\mathbf{x}_{1:t'},\mathbf{y}_{1:t'}\}$,
Preferences: $P^{\omega_{i}}=\{(\mathbf{x}\succ\mathbf{x}')_{1:p}\:|\:\forall i\in\mathbb{N}_{m}\}$
\begin{enumerate}
\item \textbf{for} $t=t'+1,\cdots,T$ iterations \textbf{do}
\item $\quad$optimize hyperparameters $\Theta_{t}^{\star}=\{\theta_{\omega_{1:m}}^{\star},\theta_{h}^{\star},\theta_{f}^{\star}\}$
and update {\footnotesize{}$\mathcal{GP}_{\omega_{1:m}},\mathcal{GP}_{h},\mathcal{GP}_{f}$} 
\item $\quad$compute predictive likelihoods $\mathcal{L}_{t}^{\mathfrak{h}}$
and $\mathcal{L}_{t}^{\mathfrak{f}}$ for Arm-$\mathfrak{h}$ and
Arm-$\mathfrak{f}$
\item $\quad$ \textbf{if} $\mathcal{L}_{t}^{\mathfrak{h}}>$ $\mathcal{L}_{t}^{\mathfrak{f}}$,
\textbf{then}
\item $\quad$$\quad$draw a random sample $S_{t}^{\mathfrak{h}}$ from
Arm-$\mathfrak{h}$ using Thompson Sampling
\item $\quad$$\quad$maximize $S_{t}^{\mathfrak{h}}$ to find $\mathbf{x}_{t}^{\mathfrak{h}}=\text{\ensuremath{\underset{\mathbf{x\in\mathcal{X}}}{\operatorname{argmax}}\,}}(S_{t}^{\mathfrak{h}}(\tilde{\mathbf{x}}))$
\item $\quad$$\quad$$\mathbf{x}_{t}=\mathbf{x}_{t}^{\mathfrak{h}}$
\item $\quad$\textbf{else},
\item $\quad$$\quad$draw a random sample $S_{t}^{\mathfrak{f}}$ Arm-$\mathfrak{f}$
using Thompson Sampling
\item $\quad$$\quad$maximize $S_{t}^{\mathfrak{f}}$ to find $\mathbf{x}_{t}^{\mathfrak{f}}=\text{\ensuremath{\underset{\mathbf{x}\in\mathcal{X}}{\operatorname{argmax}}\,}}(S_{t}^{\mathfrak{f}}(\mathbf{x}))$
\item $\quad$$\quad$$\mathbf{x}_{t}=\mathbf{x}_{t}^{\mathfrak{f}}$
\item $\quad$evaluate $f$ at $\mathbf{x}_{t}$ to obtain $y_{t}=f(\mathbf{x}_{t})+\eta_{t}$
\item $\quad$Augment data $\mathcal{D}=\mathcal{D}\cup(\mathbf{x}_{t},y_{t})$
and update expert preferences $P^{\omega_{1:m}}$ with respect to
$\mathbf{x}_{t}$ 
\item $\quad$$\mathbf{x}^{\star}=\underset{(\mathbf{x},y)\in\mathcal{D}}{\operatorname{argmax}\:y}$
\item \textbf{end for}
\item return $\mathbf{x}^{\star}$

\end{enumerate}

\end{algorithm}

\section{Convergence Remarks\label{sec:convergence}}
In this section we discuss the convergence properties of BOAP algorithm.  The essence of BOAP 
is the combination of Thompson-Sampling Bayesian Optimization (TS-BO) with 
likelihood-based model-selection from multiple models of the same objective.  To 
understand the convergence properties, we begin with special cases and then 
generalize to the real case.

{\bf Purely Un-Augmented Case:} Let us suppose that the non-augmented 
(Control arm) model is used for all iterations of BOAP.  In this mode of operation 
BOAP will operate precisely like standard Thompson-sampling BO (TS-BO), and 
thus regret is bounded as per TS-BO \cite{Rus2,Kan3,Cho7} - i.e., in 
this simplified case the (cumulative) regret is bounded as $R_T\sim\mathcal{O} 
(\sqrt{T} (B \sqrt{\gamma_T}+\gamma_T))$ if we assume $f\in\mathcal{H}_K$ 
and $\|B\|_{\mathcal{H}_K} \leq B$, where $\gamma_T$ is the maximum information 
gain (which is controlled by the covariance prior $k$).  Alternatively, 
assuming $f \in \mathcal{F}$, the Bayes regret is bounded as ${\rm 
BR}_T \sim \mathcal{O} (\sqrt{{\rm dim}_E (\mathcal{F},\frac{1}{T})T\log T})$ 
\cite{Rus2}, where ${\rm dim}_E (\mathcal{F},\frac{1}{T})$ is the eluder 
dimension of $\mathcal{F}$ \cite{Rus2,Li12} as defined below (see Definition 3.2).
\begin{definition}[$\epsilon$-independence] \cite{Rus2}
 Let $\mathcal{F}\subset\mathbb{R}^{\mathcal{X}}$, $\epsilon\in\mathbb{R}^+$.  
 Then ${\bf x} \in \mathcal{X}$ is {\em $\epsilon$-dependent of $\{ {\bf x}_1, 
 {\bf x}_2, \ldots, {\bf x}_n \} \subset \mathcal{X}$} if, $\forall f,f' \in 
 \mathcal{F}$ such that $\sum_i ( f ({\bf x}_i) - f' ({\bf x}_i))^2 \leq 
 \epsilon^2$, then $| f({\bf x}) - f'({\bf x}) | < \epsilon$.  We say that 
 ${\bf x} \in \mathcal{X}$ is {\em $\epsilon$-independent} if it is not 
 $\epsilon$-dependent.
\end{definition}
\begin{definition}[Eluder dimension] \cite{Rus2}
 Let $\mathcal{F}\subset\mathbb{R}^{\mathcal{X}}$, $\epsilon\in\mathbb{R}^+$.  
 The {\em eluder dimension ${\rm dim}_E(\mathcal{F}, \epsilon)$} of 
 $\mathcal{F}$ is the length of the longest sequence $\{ {\bf x}_1, {\bf 
 x}_2, \ldots, {\bf x}_n \} \subset \mathcal{X}$ such that $\forall i$, 
 $\exists \epsilon' \geq \epsilon$ such that ${\bf x}_i$ is 
 $\epsilon'$-independent of $\{ {\bf x}_1,{\bf x}_2,\ldots,{\bf x}_{i-1}\}$.
\end{definition}
Eluder Dimension is a measure of effective dimension, being the number of observations 
required to model any function in the set to within a given accuracy.

{\bf Purely Augmented Case:} Next, suppose that the augmented model (Human Arm) is used 
for all iterations of BOAP.  In this case the overall objective model is not 
a GP (due to the involvement of rank GP as an input to the main GP), so we cannot naively apply information-gain based TS-BO regret bounds 
as we may for the Control Arm case. However we can apply the eluder 
dimension bounds.  Whether this provides better convergence depends entirely 
on the relevance and accuracy of the user expert feedback used to construct 
the model augmentation: accurate feedback of relevant abstract properties 
should, we postulate, reduce the eluder dimension of the model (with the 
limiting case where the augmentation models $f$), while inaccurate or 
irrelevant feedback may mislead the model and increase eluder dimension, 
impeding convergence.

{\bf General Case:} More generally, BOAP may select between the Control and 
Experimental Arms, which can be modeled stochastically.  If 
neither model has a consistently higher likelihood (either initially or 
asymptotically) then the convergence behaviour 
will follow the worst-case convergence of TS-BO using either the un-augmented 
or augmented model alone - that is:
\[
 \begin{array}{rl}
  {\rm BR}_T \sim \mathcal{O} (\sqrt{\max \{{\rm dim}_E (\mathcal{F}_{\mathfrak{f}},\frac{1}{T}), {\rm dim}_E (\mathcal{F}_{\mathfrak{h}},\frac{1}{T})\}T\log T})
 \end{array}
\]
Under the reasonable assumptions that the predictive likelihood 
is (a) an accurate measure of model fit, and that asymptotically (b) 
$\mathcal{L}_t^{\mathfrak{h}} > \mathcal{L}_t^{\mathfrak{f}}$ if 
${\rm dim}_E (\mathcal{F}_{\mathfrak{h}},\frac{1}{T}) < {\rm dim}_E (\mathcal{F}_{\mathfrak{f}},\frac{1}{T})$ 
(and vice-versa) then:
\[
 \begin{array}{rl}
  {\rm BR}_T \sim \mathcal{O} (\sqrt{\min \{{\rm dim}_E (\mathcal{F}_{\mathfrak{f}},\frac{1}{T}), {\rm dim}_E (\mathcal{F}_{\mathfrak{h}},\frac{1}{T})\}T\log T})
 \end{array}
\]
So, provided that the expert provides accurate and relevant 
preference feedback we would have that (a) the augmented model will (asymptotically) be 
selected and thus dominate the regret bound, and (b) due to its lower 
eluder dimension the regret bound will be tighter, leading to a faster 
convergence.

\section{Experiments\label{sec:Experiments}}
We evaluate the performance of BOAP method using synthetic benchmark
function optimization problems and real-world optimization problems
arising in advanced battery manufacturing processes. We have considered
the following experimental settings for BOAP. We use the popular Automatic
Relevance Determination (ARD) kernel \cite{neal2012bayesian} for
the construction of both the rank GPs and the conventional (un-augmented)
GPs. For rank GPs, we tune ARD kernel hyperparameters $\theta_{d}=\{l_{d}\}$\emph{
}using max-likelihood estimation (Eq. (\ref{eq:closed_gp_log_likelihood})).
For the augmented GP modeling $\tilde{\mathbf{x}}$, we use a modified
ARD kernel that uses spatially varying kernel with a parametric lengthscale
function (See discussion in $\S$ \ref{subsec:Augmented-GP}). As
we normalize the bounds, we tune $l_{d}$ (the lengthscale for the
\textit{un}-augmented features) in the interval $[0.1,1]$ and the
scale parameter $\alpha$ (for the auxiliary features) in the interval
$(0,2]$. Further, we set signal variance $\sigma_{f}^{2}=1$ as we
standardize the outputs. 

We compare the performance of BOAP algorithm with the following state-of-the-art
baselines.\textbf{ (i) BO-TS: }a standard Bayesian Optimization (BO)
with Thompson Sampling (TS) strategy, \textbf{(ii) BO-EI: }BO with
Expected Improvement (EI) acquisition function, and \textbf{(iii)}
BOAP - Only Augmentation \textbf{(BOAP-OA)}: In this baseline, we
run our algorithm without the 2-arm scheme and we only use augmented
input for GP modeling. This baseline shows the effectiveness of expert's
inputs.. We do not consider any preference based BO methods \cite{gonzalez2017preferential,benavoli2021preferential}
as baselines, because the preferences are provided directly on the
objective function, as opposed to abstract properties that are not
measured directly. The additional experimental results and details
of our experimental setup are provided in the Appendix ($\S$ A.4.1). 

\subsection{Synthetic Experiments}

We evaluate BOAP framework in the global optimization of synthetic
optimization benchmark functions \cite{sfu_simulationlib}. The list
of synthetic functions used are provided in Table \ref{tab:syn_func_details}.

\paragraph{Emulating Preferential Expert Inputs:}

As discussed in $\S$\ref{subsec:Expert-Preferential-Inputs}, we
fit a rank GP using the expert preferences provided on designs based
on their cognitive knowledge. In all our synthetic experiments we
set $m=2$,\emph{ i.e., }we model two abstract properties $\{\omega_{1},\omega_{2}\}$
for the considered synthetic function. We expect the expert to know
the higher order abstract features of each design $\mathbf{x}\in\mathcal{X}$.
We construct rank GPs by emulating the expert preferences based on
such high-level features of the given synthetic function. The possible
set of high-level features of the synthetic functions are mentioned
in Table \ref{tab:syn_func_details}. We generate preference list
$P^{\omega_{1:2}}$ for each high-level feature of the designs by
comparing its utility. We start with $p=\binom{t'}{2}$ preferences
in $P$, that gets updated in every iteration of the optimization
process. We construct rank GP surrogates $\{\mathcal{GP}_{\omega_{1}},\mathcal{GP}_{\omega_{2}}\}$
using $P^{\omega_{1}}$ and $P^{\omega_{2}}$.

\begin{table*}
\centering{}\caption{Details of the synthetic optimization benchmark functions. Analytical
forms are provided in the $2^{\text{nd}}$ column and the last column
depicts the high level features used by a simulated expert.}

\label{tab:syn_func_details}{\small{}}%
\begin{tabular}{|l|l|l|}
\hline
{\footnotesize{}Functions} & $f(\mathbf{x})$ & High Level Features\tabularnewline
\hline
{\footnotesize{}Benchmark-$1d$} & $\exp^{(2-\mathbf{x})^{2}}+\exp^{\frac{(6-\mathbf{x})^{2}}{10}}+\frac{1}{\mathbf{x}^{2}+1}$ & $\omega_{1}=\exp^{(2-\mathbf{x})^{2}}$, $\omega_{2}=\frac{1}{\mathbf{x}^{2}}$\tabularnewline
\hline
\multirow{2}{*}{{\footnotesize{}Rosenbrock-$3d$}} & \multirow{2}{*}{$\sum\limits _{i=1}^{d-1}[100\times(x_{i+1}-x_{i}^{2})^{2}+(x_{i}-1)^{2}]$} & $\omega_{1}=(x_{3}-x_{2}^{2})^{2}+(x_{2}-x_{1}^{2})^{2}$\tabularnewline
 &  & $\omega_{2}=(x_{2}-1)^{2}+(x_{1}-1)^{2}$\tabularnewline
\hline
{\footnotesize{}Griewank-$5d$} & $\sum\limits _{i=1}^{d}\bigg[\frac{x_{i}^{2}}{4000}-\prod\limits _{i=1}^{d}\cos\big(\frac{x_{i}}{\sqrt{i}}\big)+1\bigg]$ & $\omega_{1}=\sum\limits _{i=1}^{d}x_{i}^{2},\;\omega_{2}=\prod\limits _{i=1}^{d}\cos x_{i}$\tabularnewline
\hline
\end{tabular}
\end{table*}

\begin{figure*}
\noindent \centering{}%
\begin{tabular}{ccc}
\includegraphics[width=0.31\textwidth]{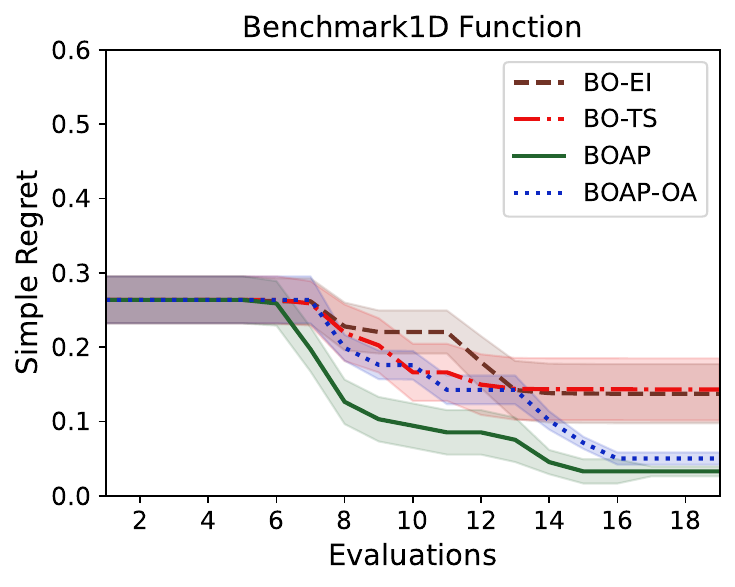} & \includegraphics[width=0.31\textwidth]{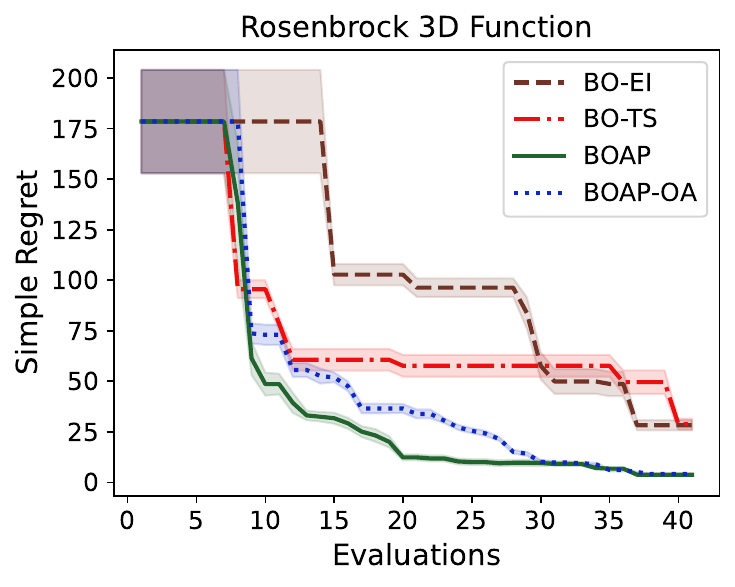} & \includegraphics[width=0.31\textwidth]{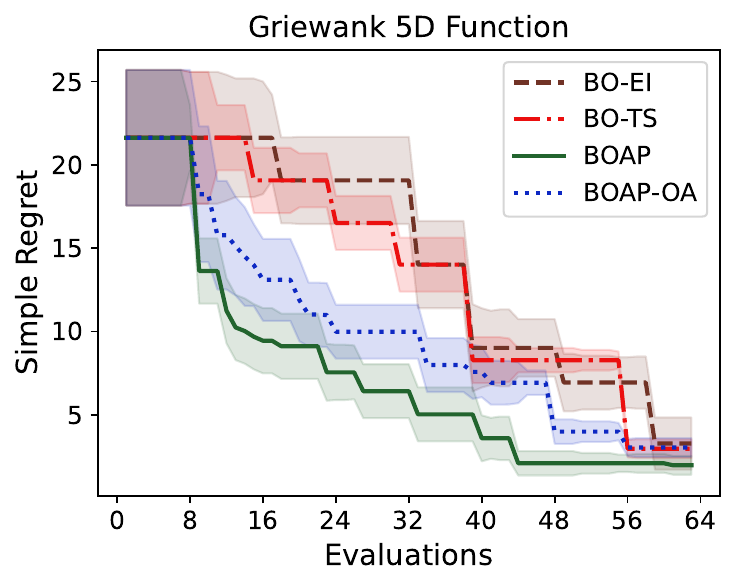}\tabularnewline
\end{tabular}\caption{Simple regret vs iterations for the synthetic multi-dimensional benchmark
functions. We plot the average regret along with its standard error
obtained after 10 random repeated runs.}
\label{fig:syn_res}
\end{figure*}

For a given $d-$dimensional problem, we have considered $t'=d+3$
initial observations and allocate the overall budget $T=10\times d+5$.
We repeat all our synthetic experiments $10$ times with random initialization
and report the average simple regret \cite{brochu2010tutorial} as
a function of iterations. The convergence plots obtained for the global
optimization experiments of synthetic functions after $10$ runs are
shown in Figure \ref{fig:syn_res}. It is evident from the convergence
results that our proposed BOAP method has outperformed the standard
baselines by a huge margin, thereby proving its superiority. 

To demonstrate the effectiveness of BOAP we have conducted additional
experiments by accounting for the inaccuracy or poor choices in expert
preferential knowledge. Please refer to Appendix
($\S$ A.4) for the additional details and experimental results.

\subsection{Real-world Experiments}

We compare the performance of BOAP in two real-world optimization
use-cases in Lithium-ion battery manufacturing that are proven to
be very complex and expensive in nature, thus providing a wide scope
for the optimization. Further, battery scientists often reveal additional
knowledge about the abstract properties in the battery design space
and thus providing a rich playground for the evaluation of our framework.
We refer to the appendix ($\S$ A.4.3) for the detailed experimental
setup.

\subsubsection{Optimization of Electrode Calendering}

In this experiment, we consider a case study on the calendering process
proposed by \cite{duq2020_calendering}. The authors analyzed the
effect of parameters such as calendering pressure ($\varepsilon_{\text{cal}}$),
electrode porosity and electrode composition on the electrode properties
such as electrolyte conductivity, tortuosity (both in solid phase
($\tau_{\text{sol}}$) and liquid phase ($\tau_{\text{liq}}$)), Current
Collector (CC), Active Surface (AS), etc. We define an optimization
paradigm using the data published by \cite{duq2020_calendering}. 

We use our proposed BOAP framework to optimize the electrode calendering
process by maximizing the \textit{Active Surface }of electrodes by
modeling two abstract properties: \textbf{(i) }Property 1 ($\omega_{1}$):
\textit{Tortuosity in liquid phase }$\tau_{\text{liq}}$, and \textbf{(ii)
}Property 2 ($\omega_{2}$): \textit{Output Porosity }(OP). We simulate
the expert pairwise preferential inputs $\{P^{\omega_{\tau_{\text{liq}}}},P^{\omega_{\text{OP}}}\}$
by comparing the actual measurements reported in the dataset published
by \cite{duq2020_calendering}. We consider $4$ initial observations
and maximize the active surface of the electrodes for $50$ iterations.
We compare the performance of our proposed BOAP framework by plotting
the average simple regret after $10$ repeated runs with random initialization.
The convergence results obtained for the electrode optimization is
shown in Figure \ref{fig:battery_calendering}.

\subsubsection{Electrode Manufacturing Optimization}

\begin{figure}
\centering{}\subfloat{\includegraphics[width=0.5\columnwidth]{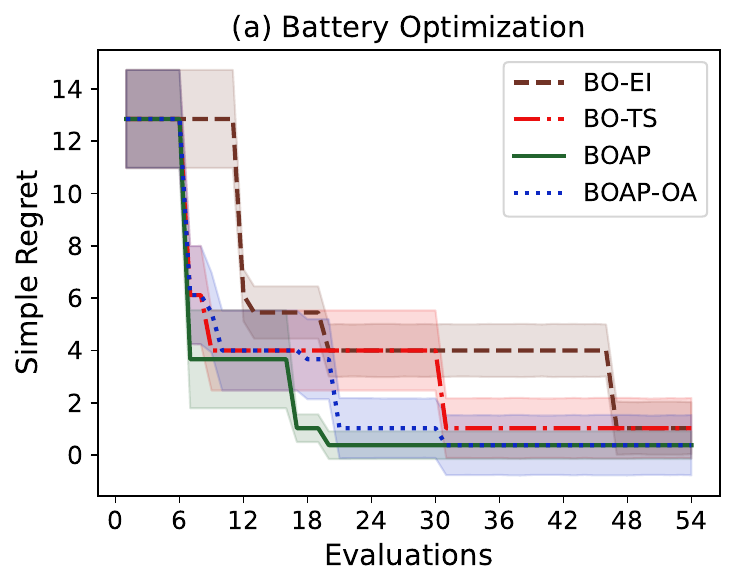}\label{fig:battery_calendering}}$\;\;$\subfloat{\includegraphics[width=0.5\columnwidth]{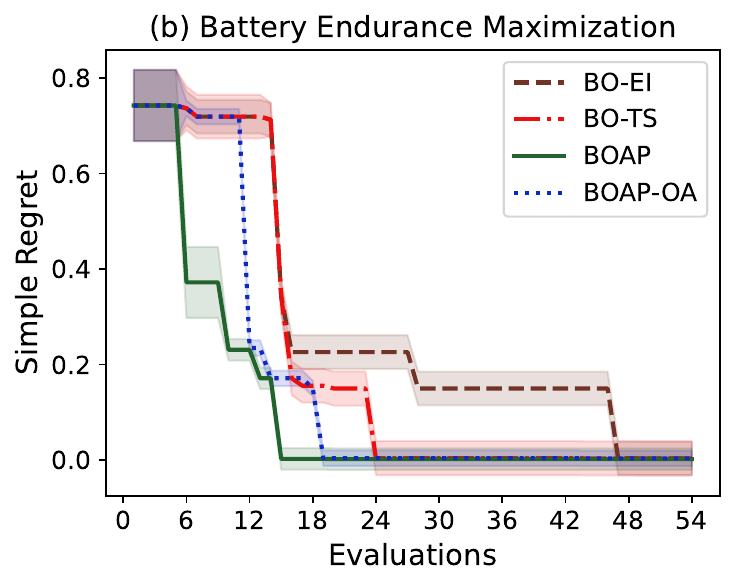}\label{fig:battery_manufacturing}}\caption{Simple regret vs iterations for battery manufacturing optimization
experiments: \textbf{(a) }Optimization of calendering process \textbf{(b)
}Optimization of battery endurance.\label{fig:realworld}\vspace{-0.2cm}
}
\end{figure}

The best battery formulation and the optimal selection of process
parameters is crucial for manufacturing long-life and energy-dense
batteries. \cite{drak_2021formulation_electrode} analyzed the manufacturing
of Lithium-ion graphite based electrodes and reported the process
parameters in manufacturing a battery along with the output charge
capacities of the battery measured after certain charge-discharge
cycles. In our experiment, we use BOAP to optimally select the manufacturing
process parameters to design a battery with maximum endurance\emph{
i.e,} a battery that can retain the maximum charge after certain charge-discharge
cycles. We consider \textit{Anode Thickness} (AT) and \textit{Active
Mass} (AM) as abstract properties $\{\omega_{\text{AT}},\omega_{\text{AM}}\}$
to maximize the battery endurance $E=\frac{D_{50}}{D_{5}}$, where
$D_{50}$ and $D_{5}$ are the discharge capacities of the cell at
$50^{\text{th}}$ and $5^{\text{th}}$ cycle, respectively. We consider
$4$ initial observations and maximize the endurance of the cell for
$50$ iterations. We compare the performance by plotting the average
simple regret versus iterations after $10$ random repeated runs.
The convergence results obtained for maximizing the endurance is shown
in Figure \ref{fig:battery_manufacturing}. 

It is evident from Figure \ref{fig:realworld} that BOAP is superior
to the baselines due to its ability to model the abstract properties
of the battery designs that can be beneficial in accelerating BO performance. 

\section{Conclusion}
We present a novel approach for human-AI collaborative BO for modeling
the expert inputs on abstract properties to further accelerate the
sample-efficiency of BO. Experts provide preferential inputs about
the abstract and unmeasurable properties. We model such preferential
inputs using rank GPs. We augment the inputs of a standard GP with
the output of such auxiliary rank GPs to learn the underlying preferences
in the instance space. We use a 2-arm strategy to overcome any futile
expert bias and encourage exploration. We discuss the convergence
of our proposed BOAP framework. The experimental results show the
superiority of our proposed BOAP algorithm.

\newpage

\bibliographystyle{unsrtnat}
\bibliography{BibFiles/PropBO,BibFiles/universal}

\begin{thebibliography}{43}
\providecommand{\natexlab}[1]{#1}
\providecommand{\url}[1]{\texttt{#1}}
\expandafter\ifx\csname urlstyle\endcsname\relax
  \providecommand{\doi}[1]{doi: #1}\else
  \providecommand{\doi}{doi: \begingroup \urlstyle{rm}\Url}\fi

\bibitem[Martinez-Cantin(2014)]{martinez2014bayesopt}
Ruben Martinez-Cantin.
\newblock {BayesOpt: a Bayesian optimization library for nonlinear
  optimization, experimental design and bandits}.
\newblock \emph{J. Mach. Learn. Res.}, 15\penalty0 (1):\penalty0 3735--3739,
  2014.

\bibitem[Greenhill et~al.(2020)Greenhill, Rana, Gupta, Vellanki, and
  Venkatesh]{greenhill2020bayesian}
Stewart Greenhill, Santu Rana, Sunil Gupta, Pratibha Vellanki, and Svetha
  Venkatesh.
\newblock {Bayesian optimization for adaptive experimental design: a review}.
\newblock \emph{IEEE access}, 8:\penalty0 13937--13948, 2020.

\bibitem[Srinivas et~al.(2012)Srinivas, Krause, Kakade, and
  Seeger]{srinivas2012information}
Niranjan Srinivas, Andreas Krause, Sham~M Kakade, and Matthias~W Seeger.
\newblock {Information-theoretic regret bounds for Gaussian process
  optimization in the bandit setting}.
\newblock \emph{IEEE Transactions on Information Theory}, 58\penalty0
  (5):\penalty0 3250--3265, 2012.

\bibitem[Chowdhury and Gopalan(2017{\natexlab{a}})]{chowdhury2017kernelized}
Sayak~Ray Chowdhury and Aditya Gopalan.
\newblock {On kernelized multi-armed bandits}.
\newblock In \emph{International Conference on Machine Learning}, pages
  844--853. PMLR, 2017{\natexlab{a}}.

\bibitem[Swersky(2017)]{swersky2017improving}
Kevin Swersky.
\newblock \emph{{Improving Bayesian optimization for machine learning using
  expert priors}}.
\newblock University of Toronto (Canada), 2017.

\bibitem[Venkatesh et~al.(2019)Venkatesh, Rana, Li, Gupta, Shilton, and
  Venkatesh]{kumar2019bayesian}
Arun Kumar~Anjanapura Venkatesh, Santu Rana, Cheng Li, Sunil Gupta, Alistair
  Shilton, and Svetha Venkatesh.
\newblock {Bayesian optimization for objective functions with varying
  smoothness}.
\newblock In \emph{Australasian Joint Conference on Artificial Intelligence},
  pages 460--472, 2019.

\bibitem[Li et~al.(2018)Li, Santu, Gupta, Nguyen, Venkatesh, Sutti, Rubin,
  Slezak, Height, Mohammed, and Gibson]{licheng2018}
Cheng Li, Rana Santu, Sunil Gupta, Vu~Nguyen, Svetha Venkatesh, Alessandra
  Sutti, David Rubin, Teo Slezak, Murray Height, Mazher Mohammed, and Ian
  Gibson.
\newblock {Accelerating experimental design by incorporating experimenter
  hunches}.
\newblock In \emph{2018 IEEE International Conference on Data Mining (ICDM)},
  pages 257--266, 2018.
\newblock \doi{10.1109/ICDM.2018.00041}.

\bibitem[Hvarfner et~al.(2022)Hvarfner, Stoll, Souza, Lindauer, Hutter, and
  Nardi]{hvarfner2022pi}
Carl Hvarfner, Danny Stoll, Artur Souza, Marius Lindauer, Frank Hutter, and
  Luigi Nardi.
\newblock {$\pi$BO: Augmenting Acquisition Functions with User Beliefs for
  Bayesian Optimization}.
\newblock \emph{arXiv preprint arXiv:2204.11051}, 2022.

\bibitem[Venkatesh et~al.(2022)Venkatesh, Rana, Shilton, and
  Venkatesh]{venkateshhuman}
Arun Kumar~Anjanapura Venkatesh, Santu Rana, Alistair Shilton, and Svetha
  Venkatesh.
\newblock {Human-AI Collaborative Bayesian optimization}.
\newblock In \emph{Advances in Neural Information Processing Systems}, 2022.

\bibitem[Nguyen et~al.(2021)Nguyen, Tay, Low, and Jaillet]{nguyen2021top}
Quoc~Phong Nguyen, Sebastian Tay, Bryan Kian~Hsiang Low, and Patrick Jaillet.
\newblock {Top-k ranking Bayesian optimization}.
\newblock In \emph{Proceedings of the AAAI Conference on Artificial
  Intelligence}, volume~35, pages 9135--9143, 2021.

\bibitem[Williams and Rasmussen(2006)]{williams2006gaussian}
Christopher~KI Williams and Carl~Edward Rasmussen.
\newblock \emph{{Gaussian processes for machine learning}}, volume~2.
\newblock MIT press Cambridge, MA, 2006.

\bibitem[Brochu et~al.(2010)Brochu, Cora, and De~Freitas]{brochu2010tutorial}
Eric Brochu, Vlad~M Cora, and Nando De~Freitas.
\newblock {A tutorial on Bayesian optimization of expensive cost functions,
  with application to active user modeling and hierarchical reinforcement
  learning}.
\newblock \emph{arXiv preprint arXiv:1012.2599}, 2010.

\bibitem[Frazier(2018)]{frazier2018tutorial}
Peter~I Frazier.
\newblock {A tutorial on Bayesian optimization}.
\newblock \emph{arXiv preprint arXiv:1807.02811}, 2018.

\bibitem[Kushner(1964)]{kushner1964new}
H.~J. Kushner.
\newblock {A new method of locating the maximum point of an arbitrary multipeak
  curve in the presence of noise}.
\newblock \emph{Journal of Basic Engineering}, 86\penalty0 (1):\penalty0
  97--106, 03 1964.
\newblock ISSN 0021-9223.
\newblock \doi{10.1115/1.3653121}.

\bibitem[Shah et~al.(2014)Shah, Wilson, and Ghahramani]{shah2014student}
Amar Shah, Andrew Wilson, and Zoubin Ghahramani.
\newblock {Student-t processes as alternatives to Gaussian processes}.
\newblock In \emph{Artificial intelligence and statistics}, pages 877--885,
  2014.

\bibitem[Zhang et~al.(2018)Zhang, Si, Hu, and Lei]{weiner_process}
Zhengxin Zhang, Xiaosheng Si, Changhua Hu, and Yaguo Lei.
\newblock {Degradation data analysis and remaining useful life estimation: A
  review on Wiener-process-based methods}.
\newblock \emph{European Journal of Operational Research}, 271\penalty0
  (3):\penalty0 775--796, 2018.

\bibitem[Mockus et~al.(1978)Mockus, Tiesis, and Zilinskas]{Moc2}
Jonas Mockus, Vytautas Tiesis, and Antanas Zilinskas.
\newblock {The application of Bayesian methods for seeking the extremum}.
\newblock In \emph{Towards Global Optimization}, volume~2, pages 117--129.
  September 1978.
\newblock ISBN 0-444-85171-2.

\bibitem[Thompson(1933)]{thompson1933likelihood}
William~R Thompson.
\newblock {On the likelihood that one unknown probability exceeds another in
  view of the evidence of two samples}.
\newblock \emph{Biometrika}, 25\penalty0 (3-4):\penalty0 285--294, 1933.

\bibitem[Kahneman and Tversky(2013)]{kahneman2013prospect}
Daniel Kahneman and Amos Tversky.
\newblock {Prospect theory: An analysis of decision under risk}.
\newblock In \emph{Handbook of the fundamentals of financial decision making:
  Part I}, pages 99--127. World Scientific, 2013.

\bibitem[Siroker and Koomen(2015)]{siroker2015b}
Dan Siroker and Pete Koomen.
\newblock \emph{{A/B testing: The most powerful way to turn clicks into
  customers}}.
\newblock John Wiley \& Sons, 2015.

\bibitem[Brusilovski et~al.(2007)Brusilovski, Kobsa, and
  Nejdl]{brusilovski2007adaptive}
Peter Brusilovski, Alfred Kobsa, and Wolfgang Nejdl.
\newblock \emph{{The adaptive web: methods and strategies of web
  personalization}}, volume 4321.
\newblock Springer Science \& Business Media, 2007.

\bibitem[Herbrich et~al.(2006)Herbrich, Minka, and
  Graepel]{herbrich2006trueskill}
Ralf Herbrich, Tom Minka, and Thore Graepel.
\newblock {TrueSkill: a Bayesian skill rating system}.
\newblock \emph{Advances in neural information processing systems}, 19, 2006.

\bibitem[Chu and Ghahramani(2005)]{chu2005preference}
Wei Chu and Zoubin Ghahramani.
\newblock {Preference learning with Gaussian processes}.
\newblock In \emph{Proceedings of the 22nd international conference on Machine
  learning}, pages 137--144, 2005.

\bibitem[Gonz{\'a}lez et~al.(2017)Gonz{\'a}lez, Dai, Damianou, and
  Lawrence]{gonzalez2017preferential}
Javier Gonz{\'a}lez, Zhenwen Dai, Andreas Damianou, and Neil~D Lawrence.
\newblock {Preferential bayesian optimization}.
\newblock In \emph{International Conference on Machine Learning}, pages
  1282--1291. PMLR, 2017.

\bibitem[Mikkola et~al.(2020)Mikkola, Todorovi{\'c}, J{\"a}rvi, Rinke, and
  Kaski]{mikkola2020projective}
Petrus Mikkola, Milica Todorovi{\'c}, Jari J{\"a}rvi, Patrick Rinke, and Samuel
  Kaski.
\newblock Projective preferential bayesian optimization.
\newblock In \emph{International Conference on Machine Learning}, pages
  6884--6892. PMLR, 2020.

\bibitem[Benavoli et~al.(2021)Benavoli, Azzimonti, and
  Piga]{benavoli2021preferential}
Alessio Benavoli, Dario Azzimonti, and Dario Piga.
\newblock {Preferential Bayesian optimization with skew Gaussian processes}.
\newblock In \emph{Proceedings of the Genetic and Evolutionary Computation
  Conference Companion}, pages 1842--1850, 2021.

\bibitem[Astudillo and Frazier(2020)]{astudillo2020multi}
Raul Astudillo and Peter Frazier.
\newblock {Multi-attribute Bayesian optimization with interactive preference
  learning}.
\newblock In \emph{International Conference on Artificial Intelligence and
  Statistics}, pages 4496--4507. PMLR, 2020.

\bibitem[Thurstone(2017)]{thurstone2017law}
Louis~L Thurstone.
\newblock {A law of comparative judgment}.
\newblock In \emph{Scaling}, pages 81--92. Routledge, 2017.

\bibitem[Russo and Van~Roy(2014)]{Rus2}
Daniel Russo and Benjamin Van~Roy.
\newblock Learning to optimize via posterior sampling.
\newblock \emph{Mathematics of Operations Research}, 39\penalty0 (4):\penalty0
  1221--1243, 2014.

\bibitem[Kandasamy et~al.(2018)Kandasamy, Krishnamurthy, Schneider, and
  P{\'o}czos]{Kan3}
Kirthevasan Kandasamy, Akshay Krishnamurthy, Jeff Schneider, and Barnab{\'a}s
  P{\'o}czos.
\newblock Parallelised bayesian optimisation via thompson sampling.
\newblock In \emph{International Conference on Artificial Intelligence and
  Statistics}, pages 133--142, 2018.

\bibitem[Chowdhury and Gopalan(2017{\natexlab{b}})]{Cho7}
Sayak~Ray Chowdhury and Aditya Gopalan.
\newblock On kernelized multi-armed bandits.
\newblock In Doina Precup and Yee~Whye Teh, editors, \emph{Proceedings of the
  34th International Conference on Machine Learning}, volume~70 of
  \emph{Proceedings of Machine Learning Research}, pages 844--853,
  International Convention Centre, Sydney, Australia, Aug 2017{\natexlab{b}}.
  PMLR.

\bibitem[Li et~al.(2021)Li, Kamath, Foster, and Srebro]{Li12}
Gene Li, Pritish Kamath, Dylan~J Foster, and Nathan Srebro.
\newblock Understanding the eluder dimension.
\newblock In \emph{Advances in Neural Information Processing Systems}, 2021.

\bibitem[Neal(2012)]{neal2012bayesian}
Radford~M Neal.
\newblock \emph{{Bayesian learning for neural networks}}, volume 118.
\newblock Springer Science \& Business Media, 2012.

\bibitem[Surjanovic and Bingham(2017)]{sfu_simulationlib}
S.~Surjanovic and D.~Bingham.
\newblock {Virtual library of simulation experiments: Test Functions and
  Datasets}, 2017.
\newblock URL \url{"http://www.sfu.ca/~ssurjano"}.
\newblock [Online; accessed 21-January-2023].

\bibitem[Duquesnoy et~al.(2020)Duquesnoy, Lombardo, Chouchane, Primo, and
  Franco]{duq2020_calendering}
Marc Duquesnoy, Teo Lombardo, Mehdi Chouchane, Emiliano~N Primo, and
  Alejandro~A Franco.
\newblock {Data-driven assessment of electrode calendering process by combining
  experimental results, in silico mesostructures generation and machine
  learning}.
\newblock \emph{Journal of Power Sources}, 480:\penalty0 229103, 2020.

\bibitem[Drakopoulos et~al.(2021)Drakopoulos, Gholamipour-Shirazi, MacDonald,
  Parini, Reynolds, Burnett, Pye, O'Regan, Wang, Whitehead,
  et~al.]{drak_2021formulation_electrode}
Stavros~X Drakopoulos, Azarmidokht Gholamipour-Shirazi, Paul MacDonald,
  Robert~C Parini, Carl~D Reynolds, David~L Burnett, Ben Pye, Kieran~B O'Regan,
  Guanmei Wang, Thomas~M Whitehead, et~al.
\newblock {Formulation and manufacturing optimization of Lithium-ion
  graphite-based electrodes via machine learning}.
\newblock \emph{Cell Reports Physical Science}, 2\penalty0 (12):\penalty0
  100683, 2021.

\bibitem[Aronszajn(1950)]{aronszajn1950theory}
Nachman Aronszajn.
\newblock {Theory of reproducing kernels}.
\newblock \emph{Transactions of the American mathematical society}, 68\penalty0
  (3):\penalty0 337--404, 1950.

\bibitem[Kaufmann et~al.(2012)Kaufmann, Korda, and Munos]{kaufmann2012thompson}
Emilie Kaufmann, Nathaniel Korda, and R{\'e}mi Munos.
\newblock {Thompson sampling: An asymptotically optimal finite-time analysis}.
\newblock In \emph{Algorithmic Learning Theory: 23rd International Conference,
  ALT 2012, Lyon, France, October 29-31, 2012. Proceedings 23}, pages 199--213.
  Springer, 2012.

\bibitem[Shahriari et~al.(2014)Shahriari, Wang, Hoffman, Bouchard-C{\^o}t{\'e},
  and de~Freitas]{shahriari2014entropy}
Bobak Shahriari, Ziyu Wang, Matthew~W Hoffman, Alexandre Bouchard-C{\^o}t{\'e},
  and Nando de~Freitas.
\newblock {An entropy search portfolio for Bayesian optimization}.
\newblock \emph{arXiv preprint arXiv:1406.4625}, 2014.

\bibitem[Bijl et~al.(2016)Bijl, Sch{\"o}n, van Wingerden, and
  Verhaegen]{bijl2016sequential}
Hildo Bijl, Thomas~B Sch{\"o}n, Jan-Willem van Wingerden, and Michel Verhaegen.
\newblock {A sequential Monte Carlo approach to Thompson sampling for Bayesian
  optimization}.
\newblock \emph{arXiv preprint arXiv:1604.00169}, 2016.

\bibitem[Schmidt and Skarstad(2001)]{li_battery_healthcare}
Craig~L Schmidt and Paul~M Skarstad.
\newblock {The future of Lithium and Lithium-ion batteries in implantable
  medical devices}.
\newblock \emph{Journal of power sources}, 97:\penalty0 742--746, 2001.

\bibitem[Brunarie et~al.(2011)Brunarie, Billard, Lansburg, and
  Belle]{telecom_Li_battery}
Joel Brunarie, Anne-Marie Billard, Stuart Lansburg, and Mathieu Belle.
\newblock {Lithium-ion (Li-ion) battery technology evolves to serve an extended
  range of telecom applications}.
\newblock In \emph{2011 IEEE 33rd International Telecommunications Energy
  Conference (INTELEC)}, pages 1--9. IEEE, 2011.

\bibitem[Harper et~al.(2019)Harper, Sommerville, Kendrick, Driscoll, Slater,
  Stolkin, Walton, Christensen, Heidrich, Lambert,
  et~al.]{ev_harper2019recycling}
Gavin Harper, Roberto Sommerville, Emma Kendrick, Laura Driscoll, Peter Slater,
  Rustam Stolkin, Allan Walton, Paul Christensen, Oliver Heidrich, Simon
  Lambert, et~al.
\newblock {Recycling Lithium-ion batteries from electric vehicles}.
\newblock \emph{nature}, 575\penalty0 (7781):\penalty0 75--86, 2019.

\end{thebibliography}

\newpage
\appendix

\section{Appendix}

\subsection{Reproducing Kernel Hilbert Spaces}

The kernel functions $k(\mathbf{x},\mathbf{x}'):\mathcal{X}\times\mathcal{X}\rightarrow\mathbb{R}$
used in Gaussian Process (GP) uniquely define an associated Reproducing
Kernel Hilbert Space (RKHS) \cite{aronszajn1950theory}. Formally:

\begin{definition}Let $\mathcal{H}_{k}$ be a Hilbert space of real
valued functions $f:\mathcal{X}\text{\ensuremath{\rightarrow}}\mathbb{R}$
on a non-empty set $\mathcal{X}$. A function $k:\mathcal{X}\times\mathcal{X}\text{\ensuremath{\rightarrow}}\mathbb{R}$
is a reproducing kernel of $\mathcal{H}_{k}$, and $\mathcal{H}_{k}$
a \textbf{Reproducing Kernel Hilbert Space} \textbf{(RKHS)}, if 
\begin{itemize}
\item $\forall\mathbf{x},\mathbf{x}'\in\mathcal{X}$, $k(\text{\ensuremath{\mathbf{x}}},\mathbf{x}')=\langle k(\text{\ensuremath{\cdot}},\mathbf{x}),k(\text{\text{\ensuremath{\cdot}}},\mathbf{x}')\rangle_{\mathcal{H}_{k}}$,
\item $k$ spans $\mathcal{H}_{k}$ i.e., $\forall\mathbf{x}\in\mathcal{X},$
$k(\text{\ensuremath{\cdot}},\mathbf{x})\in\mathcal{H}_{k}$,
\item $\forall\mathbf{x}\in\mathcal{X},\forall f\in\mathcal{H}_{k},\langle f(\cdot),k(\cdot,\mathbf{x})\rangle_{\mathcal{H}_{k}}=f(\mathbf{x})$
(the reproducing property).
\end{itemize}
\end{definition}\textit{ }

There exists varieties of kernels that can be used in fitting a GP
surrogate model. A kernel that depends only on the distance between
two given points i.e., $k=k(\mathbf{x}-\mathbf{x}'$) is called as
stationary kernels. Stationary kernels are also called as translation-invariant
kernels. Some of the popular kernel functions are listed below.

\subsubsection{Mat\'{e}rn Kernel }

Mat\'{e}rn kernel is a kernel that is commonly used in numerous machine
learning applications. There are two variants of Mat\'{e}rn kernels
that differ in their smoothness coefficient ($\nu$) as shown below.
\begin{equation}
k_{\text{MAT}}(\mathbf{x},\mathbf{x}')_{\nu=\frac{3}{2}}=\bigg(1+\frac{\sqrt{3}}{l}\Vert\mathbf{x}-\mathbf{x}'\Vert\bigg)\:\exp\bigg(-\frac{\sqrt{3}}{l}\Vert\mathbf{x}-\mathbf{x}'\Vert\bigg)\label{eq:matern_nu32_vanilla}
\end{equation}

\begin{equation}
k_{\text{MAT}}(\mathbf{x},\mathbf{x}')_{\nu=\frac{5}{2}}=\bigg(1+\frac{\sqrt{5}}{l}\Vert\mathbf{x}-\mathbf{x}'\Vert+\frac{\sqrt{5}}{3l^{2}}\Vert\mathbf{x}-\mathbf{x}'\Vert^{2}\bigg)\:\exp\bigg(-\frac{\sqrt{5}}{l}\Vert\mathbf{x}-\mathbf{x}'\Vert\bigg)\label{eq:matern_nu52_vanilla}
\end{equation}

where $l$ is the lengthscale hyperparameter of Mat\'{e}rn kernel. 

\subsubsection{Squared Exponential Kernel }

Squared Exponential (SE) kernel or Radial Basis Function (RBF) or
Gaussian kernel function is the popular stationary kernel function.
The closed-form formulation of the SE kernel is represented as shown
below.

\begin{equation}
k_{\text{SE}}(\mathbf{x},\mathbf{x}')=\sigma_{f}^{2}\:\exp\bigg(-\frac{1}{2l^{2}}\Vert\mathbf{x}-\mathbf{x}'\Vert^{2}\bigg)\label{eq:se_kernel}
\end{equation}
where, $\sigma_{f}^{2}$ and $l$ corresponds to the signal variance
and lengthscale hyperparameter, respectively, collectively represented
as $\Theta_{k}=\{l,\sigma_{f}^{2}\}$. 

\subsubsection{Linear Kernel }

Linear kernel is a commonly used non-stationary kernels defined as
the inner product of the input data points. The mathematical formulation
of a linear kernel is given by: 

\begin{equation}
k_{\text{LIN}}(\mathbf{x},\mathbf{x}')=\mathbf{x}'\mathbf{x}^{\intercal}+c\label{eq:linear_vanilla_bgnd}
\end{equation}
where $c$ is the bias hyperparameter of linear kernels. 

\subsubsection{Multi-kernel Learning }

Multi-kernel is a non-stationary kernel function defined as a linear
combination of stationary and non-stationary kernels. For instance,
multi-kernels can be constructed as: 

\begin{equation}
k_{\text{MKL}}(\mathbf{x},\mathbf{x}')=w_{1}\;k_{\text{SE}}(\mathbf{x},\mathbf{x}')+w_{2}\;k_{\text{MAT}}(\mathbf{x},\mathbf{x}')+w_{3}\;k_{\text{LIN}}(\mathbf{x},\mathbf{x}')\label{eq:mkl_vanilla_bgnd}
\end{equation}
 where $\mathbf{w}=[w_{1}\:w_{2}\:w_{3}]$ corresponds to the kernel
weights, that are usually tuned by maximizing GP log-likelihood. 

\subsection{Bayesian Optimization\label{subsec:Bayesian-Optimisation}}

The central idea of Bayesian Optimization (BO) strategy is to define
a prior distribution over all the possible set of objective functions
and then refine the model sequentially with the observed samples.
BO is built on top of the Bayes theorem that incorporate prior belief
about the black-box objective function under consideration. According
to Bayes theorem, given a model $\mathcal{M}$ and data $\mathcal{D}$,
the posterior probability of the model conditioned on data \textit{i.e.,}
$\mathcal{P}(\mathcal{M}|\mathcal{D})$ is directly proportional to
the likelihood of data $\mathcal{D}$ conditioned on model $\mathcal{M}$\textit{
i.e.,} $\mathcal{P}(\mathcal{D}|\mathcal{M})$, multiplied by the
prior probability of model $\mathcal{P}(\mathcal{M})$,
\begin{equation}
\mathcal{P}(\mathcal{M}|\mathcal{D})\;\propto\;\mathcal{P}(\mathcal{D}|\mathcal{M})\:\mathcal{P}(\mathcal{M})
\end{equation}

The observation model of BO is collected as $\mathcal{D}_{1:t}=\{\mathbf{x}_{1:t},\mathbf{y}_{1:t}\}$,
where $y_{t}=f(\mathbf{x}_{t})+\eta_{_{t}}$ is a noisy observation
of the black-box objective function $f$ evaluated at input location
$\mathbf{x}_{t}$ corrupted with a white Gaussian noise $\eta_{_{t}}\sim\mathcal{GP}(0,\sigma_{\eta}^{2})$.
In BO, we compute the posterior distribution $\mathcal{P}(f\,|\mathcal{D})$
by combining the prior $\mathcal{P}(f)$ with the likelihood $\mathcal{P}(\mathcal{D}|\,f)$
represented as, 
\begin{equation}
\mathcal{P}(f\,|\mathcal{D})\;\propto\;\mathcal{P}(\mathcal{D}|\,f)\:\mathcal{P}(f)
\end{equation}

The posterior distribution $\mathcal{P}(f\,|\mathcal{D})$ computed
captures our updated belief about the black-box objective function.
BO can be perceived as a two step sequential strategy. First step
focuses on defining the prior distribution that capture our prior
beliefs. Usually GPs are used in placing prior distributions. Second
step focuses on determining the best candidate that can be evaluated
next. Acquisition functions are used to find the next candidate point
with the high promise of finding the optima. An algorithm for the
standard Bayesian optimization procedure is provided in Algorithm
\ref{alg:bay_opt}.

\begin{algorithm}[H]
\caption{Standard Bayesian Optimization}
\label{alg:bay_opt}

\textbf{Input}: Set of observations $\mathcal{D}_{1:t'}=\{\mathbf{x}_{1:t'},\mathbf{y}_{1:t'}\}$,
Sampling budget $T$
\begin{enumerate}
\item \textbf{for} $t=t',\dots,T$ iterations \textbf{do}
\item $\qquad$optimize $\Theta^{\star}=\underset{\Theta}{\operatorname{argmax}}\,\log\:\mathcal{L}$
\item $\qquad$update GP model with optimal kernel hyperparameters $\Theta^{\star}$
\item $\qquad$find the next query point $\mathbf{x}_{t+1}=\underset{\textbf{\ensuremath{\mathbf{x}}}\in\mathbb{\mathcal{X}}}{\operatorname{argmax}}\,u(\mathbf{x})$
\item $\qquad$query $f(\mathbf{x})$ at $\mathbf{x}_{t+1}$ as $y_{t+1}=f(\mathbf{x}_{t+1})+\eta_{t+1}$
\item $\qquad$augment the data as $\mathcal{D}_{1:t+1}=\mathcal{D}_{1:t}\cup(\mathbf{x}_{t+1},y_{t+1})$
\item $\qquad$update GP model
\item \textbf{end for}
\end{enumerate}
\end{algorithm}

\subsubsection{Acquisition Functions}

The acquisition function guides the optimization by balancing the
trade-off between exploration and exploitation. \cite{kushner1964new}
proposed Expected Improvement (EI) acquisition function to guide the
search by taking into account both the probability and magnitude of
improvement over the current known optima. The next candidate point
is obtained by maximizing the acquisition function given as: 
\begin{equation}
u_{\text{EI}}(\mathbf{x})=\left\{ \begin{aligned}(\mu(\mathbf{x})-f(\mathbf{x}^{+}))\;\Phi(Z)+\sigma(\mathbf{x})\;\phi(Z)\qquad & \text{if}\;\sigma(\mathbf{x})>0\\
0\qquad\qquad\qquad\qquad\qquad\qquad\quad & \text{if}\;\sigma(\mathbf{x})=0
\end{aligned}
\right.\label{eq:acq-ei}
\end{equation}

\[
\mathcal{Z}=\frac{\mu(\mathbf{x})-f(\mathbf{x}^{+})}{\sigma(\mathbf{x})}
\]

where $\Phi(\mathcal{Z})$ and $\phi(\mathcal{Z})$ represents the
Cumulative Distribution Function (CDF) and Probability Distribution
Function (PDF) of the standard normal distribution, respectively and
$f(\mathbf{x}^{+})$ is the best value observed so far in the optimization. 

Gaussian Process-Upper Confidence Bound (GP-UCB) acquisition function
is another popular acquisition function defined based on confidence
bounds criteria. GP-UCB acquisition function is given as: 
\[
u_{\text{GP-UCB}}(\mathbf{x})=\mu(\mathbf{x})+\sqrt{\beta_{t}}\:\sigma(\mathbf{x})
\]
where $\beta_{t}$ is a hyperparameter that balances the exploration-exploitation
at iteration $t$. \cite{srinivas2012information} discussed in detail
the possible values for trade-off parameter $\beta_{t}$. Following
\cite{brochu2010tutorial}, we set the value for trade-off parameter
($\beta_{t}$) at iteration $t$ as:
\[
\beta_{t}=2\log\left(\frac{t^{\frac{d}{2}+2}\pi^{2}}{3\delta'}\right)
\]
where $\delta'\in(0,1)$, $d$ is the number of input dimensions. 

\subsubsection{Thompson Sampling based Bayesian Optimization\label{subsec:Thompson-Sampling-Bayesian}}

Thompson sampling \cite{thompson1933likelihood} is a randomized selection
strategy to select the next candidate for function evaluation by maximizing
a random sample drawn from the posterior distribution. There have
been significant advancements \cite{kaufmann2012thompson,shahriari2014entropy,bijl2016sequential,chowdhury2017kernelized}
in Thompson sampling literature that demonstrate the theoretical guarantees
of Thompson sampling. \cite{Rus2} provided a Bayesian regret bound
for Thompson sampling using the notion of eluder dimension \cite{Li12}.
A complete algorithm of Thompson sampling based Bayesian optimization
is provided in Algorithm \ref{alg:ts-bo}.

\begin{algorithm}[H]
\caption{Thompson Sampling based Bayesian Optimization}
\label{alg:ts-bo}

\textbf{Input}: Set of observations $\mathcal{D}_{1:t'}=\{\mathbf{x}_{1:t'},\mathbf{y}_{1:t'}\}$,
Sampling budget $T$
\begin{enumerate}
\item \textbf{for} $t=t',\cdots,T$ iterations \textbf{do}
\item $\qquad$optimize $\Theta^{*}=\underset{\Theta}{\operatorname{argmax}}\,\log\:\mathcal{L}$
\item $\qquad$update GP model with $\Theta^{*}$
\item $\qquad$draw a random sample $\mathfrak{g}_{t+1}$ from the updated
GP.
\item $\qquad$find the next query point $\mathbf{x}_{t+1}=\underset{\mathbf{x}\in\mathcal{X}}{\operatorname{argmax}}\,\mathfrak{g}_{t+1}(\mathbf{x})$
\item $\qquad$query $f(\mathbf{x})$ at $\mathbf{x}_{t+1}$ as $y_{t+1}=f(\mathbf{x}_{t+1})+\eta_{t+1}$
\item $\qquad$augment the data as $\mathcal{D}_{1:t+1}=\mathcal{D}_{1:t}\cup(\mathbf{x}_{t+1},y_{t+1})$
\item $\qquad$update GP posterior model
\item \textbf{end for}
\end{enumerate}
\end{algorithm}

At each iteration $t$+1, Thompson sampling strategy selects a point
$\mathbf{x}_{t}$ that is highly likely to be the optimum according
to the posterior distribution i.e., $\mathbf{x}_{t}$ is drawn from
the posterior distribution $\mathcal{P}_{\mathbf{x}^{\star}}(\cdot|\mathcal{D}_{1:t-1})$
conditioned on data $\mathcal{D}_{1:t-1}$. Thompson sampling strategy
simplifies if Gaussian Processes (GPs) are used for prior and posterior
distributions. For GPs, if $\mathfrak{g}$ is a sample drawn from
$\mathcal{GP}(\mu_{\mathcal{D}_{1:t-1}},k_{\mathcal{D}_{1:t-1}})$
we have:

\[
\mathcal{P}_{\mathbf{x}^{\star}}(\mathbf{x}|\mathcal{D}_{1:t-1})=\int\mathcal{P}_{\mathbf{x}^{\star}}(\mathbf{x}|\mathfrak{g})\mathcal{P}(\mathfrak{g}|\mathcal{D}_{1:t-1})\;d\mathfrak{g}
\]

The probability $\mathcal{P}_{\mathbf{x}^{\star}}(\mathbf{x}|\mathfrak{g})$
has its mass at the maximizer:
\begin{equation}
\underset{\mathbf{x}\in\mathcal{X}}{\operatorname{argmax}}\,\mathfrak{g}(\mathbf{x})\label{eq:mass_max}
\end{equation}
Using this utility, at each iteration $t+1$, we draw a sample $\mathfrak{g}_{t+1}$
from $\mathcal{GP}(\mu_{\mathcal{D}_{1:t-1}},k_{\mathcal{D}_{1:t-1}})$
and then find its maxima as per Eq. (\ref{eq:mass_max}). The obtained
maxima is used as the next candidate for function evaluation.

\subsection{Additional Details of BOAP Framework\label{subsec:Additional-Details}}

\begin{figure}[!t]
\noindent \centering{}\includegraphics[width=0.9\textwidth]{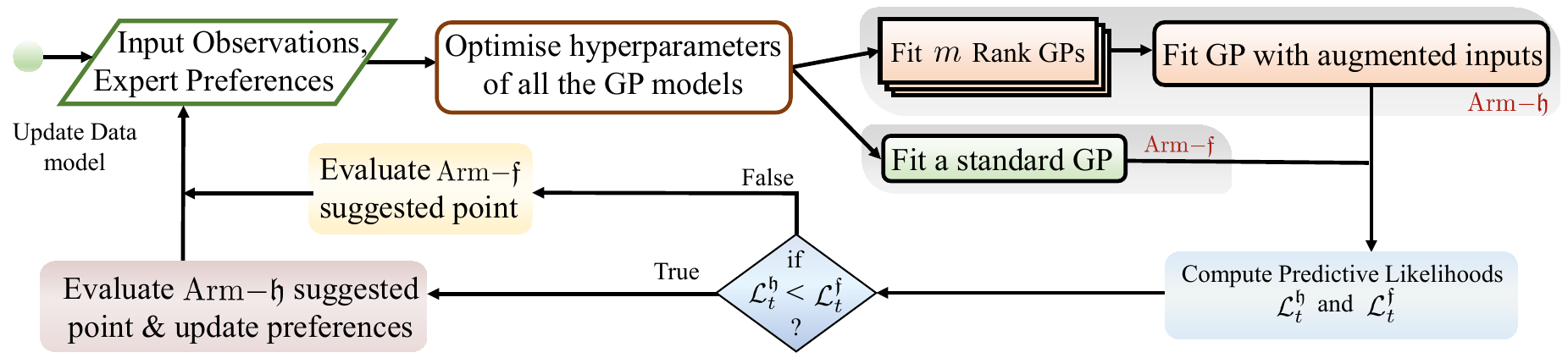}\caption{A complete process flowchart of our proposed BOAP framework.}
\label{fig:flowchart}
\end{figure}

As discussed in the main paper, our proposed framework uses a model
selection based decision making on whether to choose augmented GP
($\mathcal{GP}_{h}$) built on the expert preferential knowledge on
abstract properties or the standard GP ($\mathcal{GP}_{h}$) for suggesting
the next candidate for function evaluation. The arm containing the
standard GP model is denoted as Arm-$\mathfrak{f}$ and the arm containing
the augmented GP model is termed as Arm-$\mathfrak{h}$. A complete
flowchart of our framework is shown in Figure \ref{fig:flowchart}.

In BOAP, Arm-$\mathfrak{f}$ directly models the given objective function
$f(\mathbf{x})$, whereas Arm-$\mathfrak{h}$ models $f(\mathbf{x})$
via a human objective function ($h(\tilde{\mathbf{x}})$) in a search
space comprising of inputs that are augmented with the latent abstract
properties computed using rank GPs. As discussed in the main paper,
the human objective function $h(\tilde{\mathbf{x}})$ incorporates
the additional expert feedback from experts, and thus more-informed
and a simplified version of $f(\mathbf{x})$. Therefore, at iteration
$t$, if Arm-$\mathfrak{h}$ is selected and $\mathbf{\tilde{x}}_{t}=[\mathbf{x}_{t},\mu_{\omega_{1}}(\mathbf{x}_{t}),\cdots,\mu_{\omega_{m}}(\mathbf{x}_{t})]$
is the candidate suggested, then we observe $y_{t}$ as:
\[
y_{t}=h([\mathbf{x}_{t},\mu_{\omega_{1}}(\mathbf{x}_{t}),\cdots,\mu_{\omega_{m}}(\mathbf{x}_{t})])\approx f(\mathbf{x}_{t})
\]

\begin{figure}
\noindent \centering{}\includegraphics[width=0.8\textwidth]{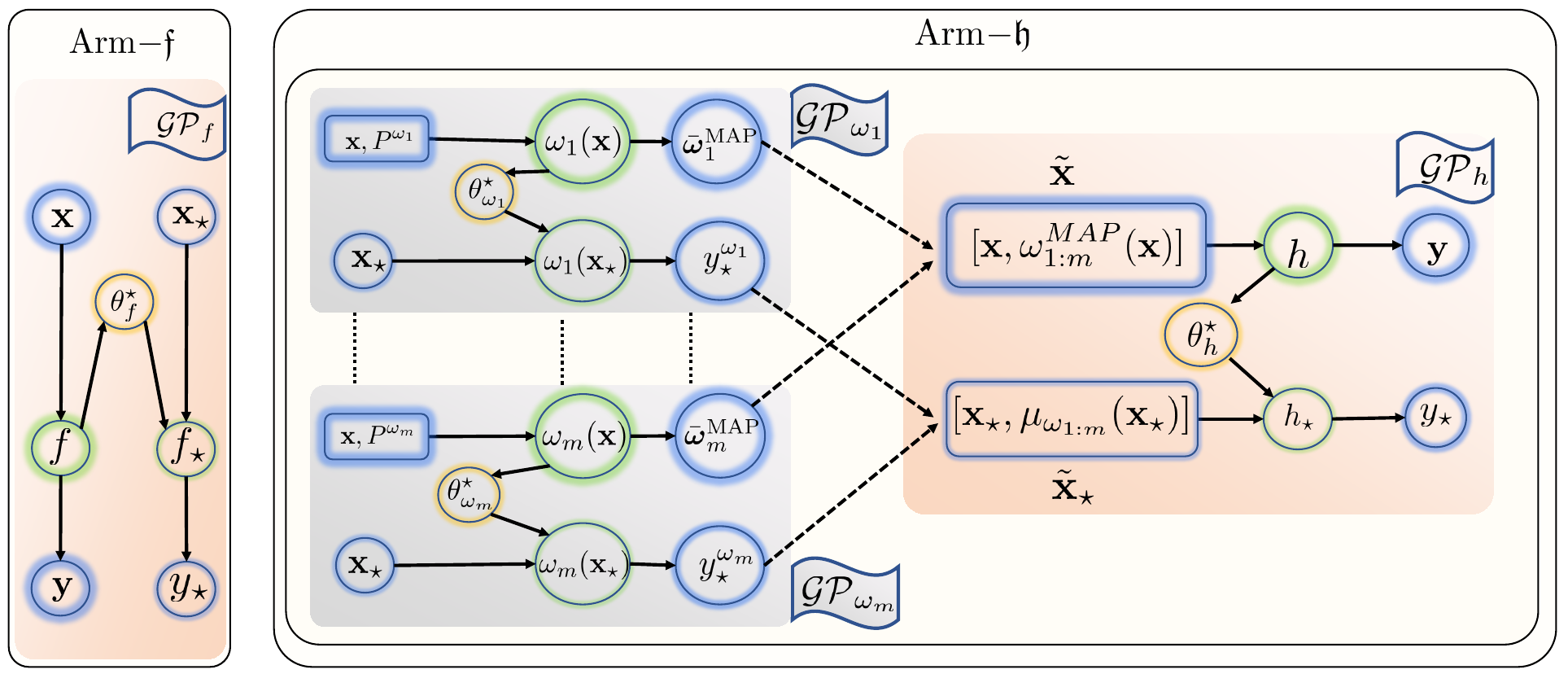}\caption{Components of Arm$\mathfrak{-f}$ and Arm$-\mathfrak{h}$ in BOAP
framework. Nodes highlighted in blue color corresponds to inputs or
outputs of a Gaussian process. The estimated parameters and the latent
variables are highlighted in yellow and green color, respectively.
Rectangular boxes shaded in Grey and Orange correspond to the nodes
representing rank GPs and conventional GPs, respectively.}
\label{fig:framework-a}
\end{figure}

A graphical representation of BOAP framework and its components are
shown in Figure \ref{fig:framework-a}. Nodes highlighted in blue
color corresponds to inputs or outputs of a Gaussian process. Rectangular
boxes shaded in Grey correspond to the nodes representing rank GPs,
whereas the rectangular boxes shaded in orange correspond to the nodes
representing conventional GPs. The estimated parameters and the latent
variables are highlighted in yellow and green color, respectively. 

\subsubsection{Derivatives of the Likelihood Function}

In rank GP distributions the MAP estimates are computed using Eq.
(4) mentioned in the main paper. However, the Newton-Raphson recursion
needs access to the first and second order derivatives of the loss
function $L=-\ln\Phi\big(z(\mathbf{x},\mathbf{x}')\big)$ with respect
to latent function values $\boldsymbol{\omega}$. The analytical formulation
of the first order and second order derivatives are given as: 

\begin{equation}
\frac{\partial L}{\partial\omega(\mathbf{x}_{i})}=\frac{\Gamma(\mathbf{x}_{i})}{\sqrt{2\sigma_{\eta}^{2}}}\frac{\phi(z)}{\Phi(z)}\label{eq:first_derv}
\end{equation}

\begin{equation}
\frac{\partial^{2}L}{\partial\omega(\mathbf{x}_{i})\partial\omega(\mathbf{x}_{j})}=\frac{\Gamma(\mathbf{x}_{i})\;\Gamma(\mathbf{x}_{j})}{2\sigma_{\eta}^{2}}\bigg(\frac{\phi^{2}(z)}{\Phi^{2}(z)}+z\frac{\phi(z)}{\Phi(z)}\bigg)\label{eq:second_derv-1}
\end{equation}
 
\[
\text{where}\qquad\Gamma_{(\mathbf{x},\mathbf{x}')}(\mathbf{x}_{i})=\left\{ \begin{aligned}1\qquad & \text{if}\;\mathbf{x}=\mathbf{x}_{i}\\
-1\qquad & \text{if}\;\mathbf{x}'=\mathbf{x}_{i}\\
0\qquad & \text{\text{Other}wise}
\end{aligned}
\right.
\]

\subsection{Additional Results and Experimental Details\label{subsec:experimental_details}}

\subsubsection{Parameter Selection in BOAP framework}

The kernel functions ($k$) used in fitting a Gaussian process model
for the unknown objective function $f$ is associated with its own
hyperparameter set $\theta$. The optimal kernel hyperparameters ($\theta^{\star}$)
are estimated by maximizing the marginal likelihood function, given
by the equation: 

\begin{equation}
\mathcal{P}(\mathbf{y}|\mathbf{X},\theta)=\int p(\mathbf{y}|f)\;p(f|\mathbf{X},\theta)\;df\label{eq:likelihood}
\end{equation}
By marginalizing Eq. (\ref{eq:likelihood}), we get the closed-form
for GP log-likelihood as:

\begin{equation}
\mathcal{L}=\log\mathcal{P}(\mathbf{y}|\mathbf{X},\theta)=-\frac{1}{2}(\mathbf{y}^{\intercal}[\mathbf{K}+\sigma_{\text{\ensuremath{\eta}}}^{2}\mathbf{I}]^{-1}\mathbf{y})-\frac{1}{2}\log|\mathbf{K}+\sigma_{\text{\ensuremath{\eta}}}^{2}\mathbf{I}|-\frac{n}{2}\log(2\pi)\label{eq:gp_log_likelihood}
\end{equation}
where $n$ corresponds to the total number of training instances.
For a rank GP the closed-form of the GP log-likelihood can be obtained
as:

\begin{equation}
\mathcal{\bar{L}}=-\frac{1}{2}\textbf{\ensuremath{\boldsymbol{\omega}_{{\scriptscriptstyle \text{MAP}}}^{\intercal}}}[\mathbf{K}\!+\sigma_{\eta}^{2}\mathbf{I}]^{{\scriptscriptstyle {\scriptscriptstyle -1}}}\textbf{\ensuremath{\boldsymbol{\omega}_{{\scriptscriptstyle \text{MAP}}}}}-\frac{1}{2}\log\mid\mathbf{K}+\sigma_{\eta}^{2}\mathbf{I}\mid-\frac{n}{2}\log(2\pi)\label{eq:rank_gp_log_likeli}
\end{equation}

The only difference in the formulation of log-likelihood of a rank
GP and a traditional GP is that the absolute measurements ($\mathbf{y}$)
of the objective function is replaced with the latent function values
obtained via Maximum A Posteriori (MAP) estimates. The GP log-likelihood
mentioned in Eq. (\ref{eq:gp_log_likelihood}) and Eq. (\ref{eq:rank_gp_log_likeli})
is now maximized to find the optimal hyperparameters $\theta^{\star}$.

\[
\theta^{\star}=\underset{\theta}{\operatorname{argmax}}\,\mathcal{L}
\]

Further, if we use all training instances for the computation of the
log marginal likelihood, there are chances that only Control arm may
get selected in majority of the rounds. Therefore to avoid this, instead
of using all the training instances for computing the marginal likelihood,
we use only the subset of the original training data for finding the
optimal hyperparameter set and then we use the held-out instances
from the original training set to compute the (predictive) likelihood. 

In this paper, we implement Automatic Relevance Determination (ARD)
kernel \cite{neal2012bayesian} based on Squared Exponential kernel
mentioned in Eq. (\ref{eq:se_kernel}) at all levels of our proposed
BOAP framework. ARD kernel is popular in the machine learning community
due to its ability to suppress the irrelevant features. In ARD, each
input dimension is assigned a different lengthscale parameter to keep
track of the relevance of that dimension. Therefore, a GP fitted in
a $d-$dimensional input space with ARD kernel has $d$ lengthscale
parameters \emph{i.e.,} $\boldsymbol{l}=l_{1:d}$ and optional variance
hyperparameters (noise variance $\sigma_{\eta}^{2}$ and signal variance
$\sigma_{f}^{2}$). 

Specifically, for the human-inspired arm (Arm-$\mathfrak{\ensuremath{h}}$)
we use a spatially-varying ARD kernel where we set the lengthscales
of the augmented input dimensions in proportion to the rank GP uncertainties
via a parametric function of the input $\mathbf{x}$. The lengthscale
function for each of the augmented input dimension (corresponding
to property $\omega_{i}$) is set to be $l_{\omega_{i}}(\mathbf{x})=\alpha\sigma_{\omega_{i}}(\mathbf{x})$,
where $\alpha$ is the scale parameter and $\sigma_{\omega_{i}}(\mathbf{x})$
is the normalized standard deviation predicted using the rank GP ($\mathcal{GP}_{\omega_{i}}$).
Therefore, the hyperparameter set ($\theta_{h}$) of Arm-$\mathfrak{\ensuremath{h}}$
consists of the lengthscale parameters ($l_{1:D}$) for the original
un-augmented dimensions and the scale factor ($\boldsymbol{\mathbf{\alpha}}$)
from the augmented input dimensions i.e., $\theta_{h}=\{l_{1:D},\boldsymbol{\mathbf{\alpha}}\}$. 

The overall hyperparameter set $\Theta$ of our proposed framework
consists of hyperparameters from each of the $m$ rank GPs ($\theta_{\omega_{1:m}}$)
and two main GPs corresponding to the 2-arms ($\theta_{h}$ and $\theta_{f}$)
\emph{i.e.,} $\Theta=\{\theta_{h},\theta_{f}\}=\{\{l_{1:D},\boldsymbol{\mathbf{\alpha}}\},\{l_{1:D}\}\}$.
At each iteration $t$, we find the optimal set of hyperparameters
$\Theta_{t}^{\star}=\{\theta_{h}^{\star},\theta_{f}^{\star}\}$ by
maximizing the GP (predictive) log-likelihood mentioned in Eq. (\ref{eq:gp_log_likelihood})
and Eq. (\ref{eq:rank_gp_log_likeli}).

In all our experiments, we set the signal variance parameter $\sigma_{f}^{2}=1$
as we standardize the outputs of $\mathcal{GP}_{h}$ and $\mathcal{GP}_{f}$.
We set the noise variance as $\eta\sim\mathcal{GP}(0,\sigma_{\eta}^{2}=0.1)$
and $\tilde{\sigma}_{\eta}^{2}=0.1$. As we normalize the input space
of the GP distributions constructed in our BOAP framework, we tune
each lengthscale hyperparameter $l\in\Theta$ in the interval $[0,1]$.
Further, we normalize the outputs of each of the auxiliary rank GPs
($\mathcal{GP}_{\omega_{i}}$) to avoid different scaling levels in
their output, that can lead to undesired structures in the augmented
input space. 

We run all our experiments on an Intel Xeon CPU@ 3.60GHz workstation
with 16 GB of RAM capacity. We repeat our experiments with $10$ different
random initialization. For a $d-$dimensional problem, we consider
$t'=d+3$ initial observations. The evaluation budget is set as $T=10\times d+5$.
For the real-world battery design experiments, due to their expensive
nature, we have restricted the evaluation budget to $50$ iterations
even though $d\gg5$. In all our experiments, we start with $p=\binom{t'}{2}$
preferences in $P$, that gets updated in every iteration of the optimization
process.

\subsubsection{Ablation Studies of BOAP Framework}

To demonstrate the robustness of our approach we have conducted additional
experiments by accounting for the inaccuracy or poor choices in expert
preferential knowledge. Here, we show the performance of our BOAP
approach in two scenarios. First, we show the performance of our proposed
approach when the higher-order abstract properties are poorly selected.
Second, we incorporate noise in the expert preferential feedback by
flipping the expert preference between two inputs (designs) with probability
$\delta=0.3$. We now discuss in detail the aforementioned two variations
of our proposed method.

\paragraph*{Inaccurate Abstract Properties (BOAP-IA)}

In the first variation, we assume that the expert poorly selects the
human abstraction features. Table \ref{tab:inacc_abs_props} depicts
the synthetic functions considered and the corresponding (poorly chosen)
human abstraction features. BOAP-IA uses such inaccurate human abstract
features while augmenting the original input space. 

\begin{table}[H]
\begin{centering}
\caption{Selection of abstraction features by a simulated human expert. The
human abstraction (high level) features shown in the 3rd column are
deliberately selected to be uninformative.}
{\small{}\label{tab:inacc_abs_props}}{\small\par}
\par\end{centering}
\centering{}{\small{}}%
\begin{tabular}{ccc}
\toprule 
Functions & $f(\mathbf{x})$ & Human Abstraction Features\tabularnewline
\midrule
Benchmark & $\exp^{(2-\mathbf{x})^{2}}+\exp^{\frac{(6-\mathbf{x})^{2}}{10}}+\frac{1}{\mathbf{x}^{2}+1}$ & $\omega_{1}=\sin\mathbf{x}$, $\omega_{2}=\cos\mathbf{x}$\tabularnewline
\midrule
\multirow{2}{*}{Rosenbrock} & \multirow{2}{*}{$\sum\limits _{i=1}^{d-1}[100\times(x_{i+1}-x_{i}^{2})^{2}+(x_{i}-1)^{2}]$} & \multirow{2}{*}{$\omega_{1}=\sin\mathbf{x}$, $\omega_{2}=\cos\mathbf{x}$}\tabularnewline
 &  & \tabularnewline
\midrule
Griewank & $\sum\limits _{i=1}^{d}\bigg[\frac{x_{i}^{2}}{4000}-\prod\limits _{i=1}^{d}\cos\big(\frac{x_{i}}{\sqrt{i}}\big)+1\bigg]$ & $\omega_{1}=\sin\mathbf{x}$, $\omega_{2}=\mathbf{x}^{3}$\tabularnewline
\bottomrule
\end{tabular}{\small\par}
\end{table}

\paragraph*{Noisy Expert Preferences (BOAP-NP)}

In the second variation, we account for the inaccurate expert preferential
knowledge by introducing an error in human expert preferential feedback.
To do this, we flip the preference ordering with a probability $\delta$
(we used $\delta=0.3$) i.e., $P^{\omega,\delta}=\{(\mathbf{x}{}_{i}\succ\mathbf{x}{}_{j})\:|\:\mathbf{x}{}_{i},\mathbf{x}{}_{j}\in\mathbf{x}{}_{1:n},\nu_{ij}\:\omega(\mathbf{x}{}_{i})>\nu_{ij}\:\omega(\mathbf{x}{}_{j})$\},
where $\nu_{ij}$ is drawn from a random distribution such that it
is $+1$ with probability $1-\delta$, $-1$ with probability $\delta$. 

\begin{figure}
\noindent \centering{}%
\begin{tabular}{ccc}
\includegraphics[width=0.31\textwidth]{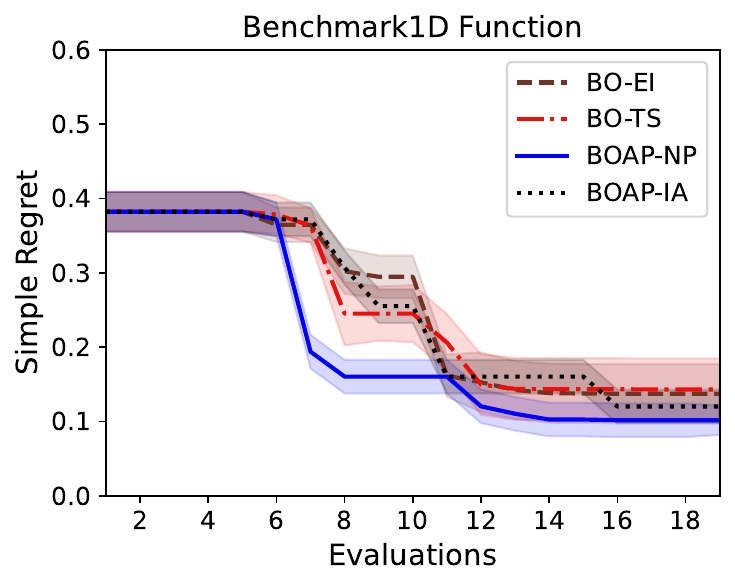} & \includegraphics[width=0.31\textwidth]{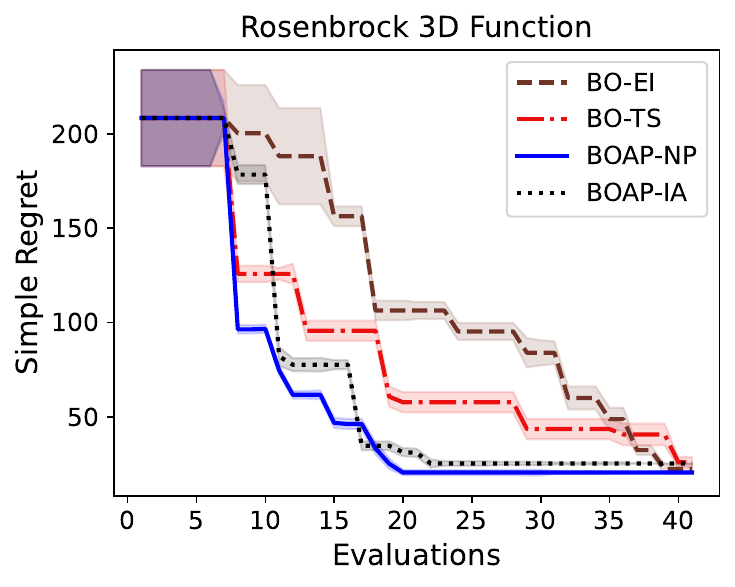} & \includegraphics[width=0.31\textwidth]{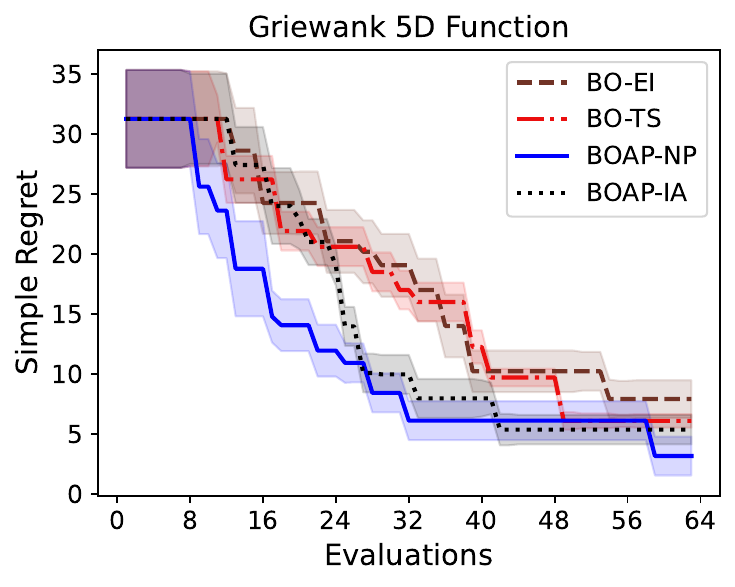}\tabularnewline
\end{tabular}\caption{Simple regret vs iterations for the synthetic multi-dimensional benchmark
functions. We plot the average regret along with its standard error
obtained after 10 random repeated runs.}
\label{fig:inacc_noisy_exp_feed}
\end{figure}

We evaluate the performance by computing the simple regret after $10d$
iterations. The experimental settings are retained as mentioned in
the experiments section (refer to Section 5.1 in the main paper).
The empirical results for BOAP with inaccurate features (BOAP-IA)
and BOAP framework with noisy preferences (BOAP-NP) are presented
in Figure \ref{fig:inacc_noisy_exp_feed}. It is significant from
the results that our proposed BOAP framework outperforms standard
baselines due to the model selection based 2-arm scheme employed that
intelligently chooses the arm with maximum predictive likelihood to
suggest the next sample. 

\subsubsection{Details of Real-world Experiments}

As mentioned in the main paper, we evaluate the performance of BOAP
framework in two real-world optimization paradigms in the Lithium-ion
battery manufacturing. The advent of cheap Li-ion battery technology
has significantly transformed range of industries including healthcare
\cite{li_battery_healthcare}, telecommunication \cite{telecom_Li_battery},
automobiles \cite{ev_harper2019recycling}, and many more due to its
ability to efficiently store the electrochemical energy. However,
the process for manufacturing of Li-ion batteries is very complex
and expensive in nature. Thus, there is a wide scope for optimizing
the battery manufacturing process to reduce its CapEx and OpEx. We
now provide a brief discussion on battery manufacturing experiments
considered in our main paper. For real-world experiments, we use the
same set of algorithmic parameters that we used in the synthetic experiments.

\paragraph{Optimization of Electrode Calendering Process\protect \\
}

\cite{duq2020_calendering} discussed the effect of calendering process
on electrode properties that significantly contribute to the underlying
electrochemical performance of a battery. Authors have implemented
a data-driven stochastic electrode structure generator, based on which
they construct electrodes and analyze in terms of \emph{Tortuosity
}(both in solid phase $\tau_{\text{sol}}$ and liquid phase $\tau_{\text{liq}}$),
percentage of \emph{Current Collector }(CC) surface covered by the
active material and percentage of \emph{Active Surface} (AS) covered
by the electrolyte. The manufacturing process parameters considered
are calendering pressure, Carbon-Binder Domain (CBD), initial electrode
porosity and electrode composition. A pictorial representation of
the inter dependencies between the input process variables and output
electrode properties is provided in Figure 8 of \cite{duq2020_calendering}.

\cite{duq2020_calendering} published a dataset reporting the input
manufacturing process parameters and the output characteristics of
$8800$ electrode structures. Each of the manufacturing setting has
been evaluated for $10$ times, therefore we have averaged the results
to obtain a refined dataset consisting of $n=880$ instances with
$d=8$ process variables.

We optimize the calendering process by maximizing the \textit{Active
Surface }of an electrode (overall objective) by modeling two abstract
properties: \textbf{(i) }Property 1 ($\omega_{1}$): \textit{Tortuosity
in liquid phase }$\tau_{\text{liq}}$, and \textbf{(ii) }Property
2 ($\omega_{2}$): \textit{Output Porosity }(OP). As discussed in
the main paper, the abstract properties $\{\omega_{\tau_{\text{liq}}},\omega_{\text{OP}}\}$
can only be qualitatively measured, however to simulate the expert
pairwise preferential inputs $\{P^{\omega_{\tau_{\text{liq}}}},P^{\omega_{\text{OP}}}\}$,
we use the empirical measurements for\textit{ }$\tau_{\text{liq}}$
and OP in the published dataset.
\[
P^{\omega_{\tau_{\text{liq}}}}=\{(\mathbf{x}\succ\mathbf{x}')_{i}\:\text{if}\:\tau_{\text{liq}}(\mathbf{x})>\tau_{\text{liq}}(\mathbf{x}')\:|\;\mathbf{x},\mathbf{x}'\in\mathbf{x}_{1:n}\;\;\forall i\in\mathbb{N}_{p}\}
\]

\[
P^{\omega_{\text{OP}}}=\{(\mathbf{x}\succ\mathbf{x}')_{i}\:\text{if}\:\text{OP}(\mathbf{x})>\text{OP}(\mathbf{x}')\:|\;\mathbf{x},\mathbf{x}'\in\mathbf{x}_{1:n}\;\;\forall i\in\mathbb{N}_{p}\}
\]

We obtain the values $\tau_{\text{liq}}(\mathbf{x})$ and $\text{OP}(\mathbf{x})$
by referring to the dataset published. Based on these preference lists
$\{P^{\omega_{\tau_{\text{liq}}}},P^{\omega_{\text{OP}}}\}$, we fit
two auxiliary rank GP distributions $\{\mathcal{GP}_{\omega_{\tau_{\text{liq}}}},\mathcal{GP}_{\omega_{\text{OP}}}\}$.
Then, we use these auxiliary rank GPs to augment the input space of
the main GP surrogate $\mathcal{GP}_{h}$ modeling the overall objective
i.e., active surface of the electrode. 

\paragraph{Optimization of Electrode Manufacturing Process\protect \\
}

In a similar case study \cite{drak_2021formulation_electrode}, the
authors have analyzed the manufacturing of Lithium-ion graphite based
electrodes. The main aim of \cite{drak_2021formulation_electrode}
is to optimize the formulation and manufacturing process of Lithium-ion
electrodes using machine learning. Authors have established a relationship
between the process parameters at different stages of manufacturing
such as mixing, coating, drying, and calendering. The published dataset
records all the process parameters in manufacturing $256$ coin cells,
as well as the associated results showing the charge capacity of each
coin cell measured after certain charge-discharge cycles. The refined
dataset consists of $12$ process variables, out of which two process
variables: \textbf{(i)} \emph{Anode Thickness }(AT), and \textbf{(ii)}\emph{
Active Mass }(AM) are treated as abstract properties that can be only
qualitatively measured. The overall objective here is to maximize
the battery endurance $E=\frac{D_{50}}{D_{5}}$, where $D_{50}$ and
$D_{5}$ are the discharge capacities of the cell at $50^{\text{th}}$
cycle and $5^{\text{th}}$ cycle, respectively. 

We simulate the expert pairwise preferential inputs $\{P^{\omega_{\text{AT}}},P^{\omega_{\text{AM}}}\}$
by comparing the empirical values recorded for these variables in
the given dataset.
\[
P^{\omega_{\text{AT}}}=\{(\mathbf{x}\succ\mathbf{x}')_{i}\:\text{if}\:\text{AT}(\mathbf{x})>\text{AT}(\mathbf{x}')\:|\;\mathbf{x},\mathbf{x}'\in\mathbf{x}_{1:n}\;\;\forall i\in\mathbb{N}_{p}\}
\]

\[
P^{\omega_{\text{\text{AM}}}}=\{(\mathbf{x}\succ\mathbf{x}')_{i}\:\text{if}\:\text{AM}(\mathbf{x})>\text{AM}(\mathbf{x}')\:|\;\mathbf{x},\mathbf{x}'\in\mathbf{x}_{1:n}\;\;\forall i\in\mathbb{N}_{p}\}
\]

We model abstract properties $\{\omega_{\text{AT}},\omega_{\text{AT}}\}$
by fitting rank GPs $\{\mathcal{GP}_{\omega_{\text{AM}}},\mathcal{GP}_{\omega_{\text{AT}}}\}$
using preference lists $\{P^{\omega_{\text{AM}}},P^{\omega_{\text{AM}}}\}$.
We use rank GPs to estimate the abstract properties and then use those
estimations to augment the input space of the main GP modeling the
overall objective i.e., maximizing battery endurance $E$. 

\subsection{Limitations}

Firstly, as discussed in Convergence Remarks section in the main paper,
if the expert preferential data is inaccurate or irrelevant, it may
mislead the model and increase eluder dimension, thus impeding convergence.
Although the ablation studies show the robustness of BOAP against
inaccurate/noisy expert preferences, our framework may suffer at least
in the initial rounds of the optimization until the Control arm identifies
and dominates to suppress the inaccurate/noisy human-inspired model
(Arm-$\mathfrak{\ensuremath{h}})$. Secondly, BOAP in its vanilla
version may not be directly scalable because of the scalability issues
of the underlying range of GPs used in the framework. One of the well-known
weaknesses of GP is that it poorly scales and suffers from a cubic
time complexity $O(n^{3})$ due to the inversion of the gram matrix
$\mathbf{K}$. This limits the scalability of GP and thus our BOAP
framework to use with large-scale datasets. In the future line of
work, we aspire to overcome these limitations by employing suitable
GP techniques that can be easily scaled.

\end{document}